\documentclass[journal]{IEEEtran}
\usepackage{graphicx}
\usepackage{amstext}
\usepackage{mathrsfs}
\usepackage{amssymb}
\usepackage{amsmath}
\usepackage{bm}
\allowdisplaybreaks[4]
\usepackage{epstopdf}
\usepackage{multicol}
\usepackage{stfloats}
\usepackage{url}
\usepackage{color}
\usepackage{enumerate}
\usepackage{breqn}
\usepackage{bbm}
\usepackage{cite}
\usepackage{caption}
\usepackage{enumitem}
\usepackage{array}
\ifCLASSINFOpdf
\else
\fi
\usepackage[ruled,vlined]{algorithm2e}
\SetKwInput{KwInput}{Input}                %
\SetKwInput{KwOutput}{Output}              %
\definecolor{rc1}{RGB}{0, 0, 0}
\newcommand{\re}[1]{{\color{rc1}#1}}

\usepackage{amsmath}
\usepackage{amssymb}
\usepackage{gensymb}
\usepackage{booktabs}
\usepackage{multirow}
\usepackage{subcaption}
\usepackage{diagbox}
\DeclareMathOperator*{\argmin}{argmin}
\usepackage{xurl}
\usepackage[colorlinks,
            linkcolor=blue,
            anchorcolor=blue,
            citecolor=blue]{hyperref}

\begin{document}

\title{Inconspicuous Adversarial Patches for Fooling Image Recognition Systems on Mobile Devices}

\author{Tao Bai,
        Jinqi Luo,
        Jun Zhao,~\IEEEmembership{Member,~IEEE}
        \thanks{This paper is supported by 1) Singapore Ministry of Education Academic Research Fund Tier 1 RG128/18, Tier 1 RG115/19, Tier 1 RT07/19, Tier 1 RT01/19, Tier 1 RG24/20, and Tier 2 MOE2019-T2-1-176, 2) NTU-WASP Joint Project, 3) Singapore NRF National Satellite of Excellence, Design Science and Technology for Secure Critical Infrastructure NSoE DeST-SCI2019-0012, 4) AI Singapore (AISG) 100 Experiments (100E) programme, and 5) NTU Project for Large Vertical Take-Off \& Landing (VTOL) Research Platform. (Corresponding author: Jun Zhao.)} 

\thanks{T. Bai, J. Luo, and J. Zhao are with the School of Computer Science and Engineering, Nanyang Technological University, Singapore 639798 (Email: bait0002@ntu.edu.sg; luoj0021@ntu.edu.sg; junzhao@ntu.edu.sg).}%
\thanks{}
\thanks{}
}

\maketitle

\begin{abstract}
Deep learning based image recognition systems have been widely deployed on mobile devices in today's world.
In recent studies, however, deep learning models are shown vulnerable to adversarial examples.
One variant of adversarial examples, called adversarial patch, draws researchers' attention due to its strong attack abilities.
Though adversarial patches achieve high attack success rates, they are easily being detected because of the visual inconsistency between the patches and the original images.
Besides, it usually requires a large amount of data for adversarial patch generation in the literature, which is computationally expensive and time-consuming.
To tackle these challenges, we propose an approach to generate inconspicuous adversarial patches with one single image.
In our approach, we first decide the patch locations based on the perceptual sensitivity of victim models, then produce adversarial patches in a coarse-to-fine way by utilizing multiple-scale generators and discriminators.
The patches are encouraged to be consistent with the background images with adversarial training while preserving strong attack abilities.
Our approach shows the strong attack abilities in white-box settings and the excellent transferability in black-box settings through extensive experiments on various models with different architectures and training methods.
Compared to other adversarial patches, our adversarial patches hold the most negligible risks to be detected and can evade human observations, which is supported by the illustrations of saliency maps and results of user evaluations.
Lastly, we show that our adversarial patches can be applied in the physical world.
\end{abstract}


\IEEEpeerreviewmaketitle

\section{Introduction}

\begin{figure}[t]
\centering

\begin{subfigure}{0.19\columnwidth}
\caption*{Original}
  \includegraphics[width=\linewidth]{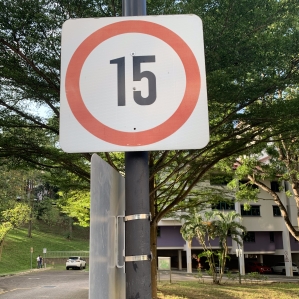}
\end{subfigure}
\begin{subfigure}{0.19\columnwidth}
\caption*{GP}
  \includegraphics[width=\linewidth]{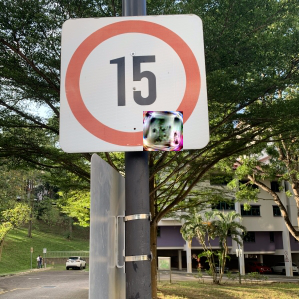}
\end{subfigure}
\begin{subfigure}{0.19\columnwidth}
\caption*{LaVAN}
  \includegraphics[width=\linewidth]{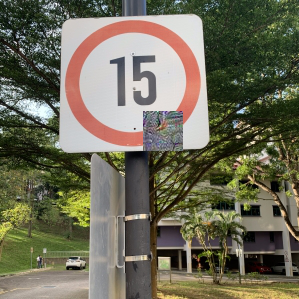}
\end{subfigure}
\begin{subfigure}{0.19\columnwidth}
\caption*{\re{PS-GAN}}
  \includegraphics[width=\linewidth]{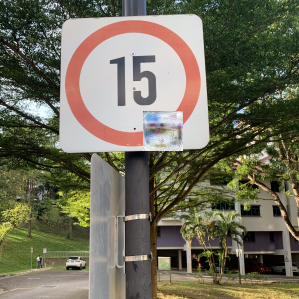}
\end{subfigure}
\begin{subfigure}{0.19\columnwidth}
\caption*{IAP (Ours)}
\includegraphics[width=\linewidth]{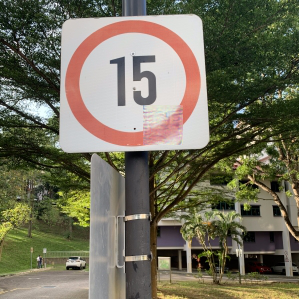}
\end{subfigure}

\vspace{.4em}

\begin{subfigure}{0.19\columnwidth}
  \includegraphics[width=\linewidth]{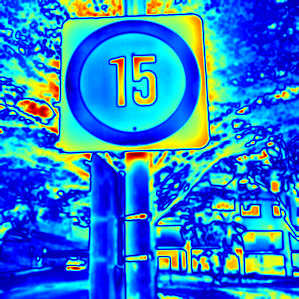}
\end{subfigure}
\begin{subfigure}{0.19\columnwidth}
  \includegraphics[width=\linewidth]{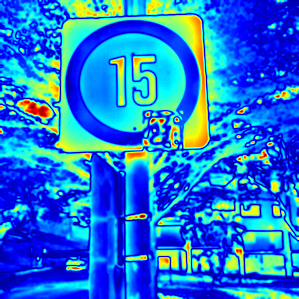}
  
\end{subfigure}
\begin{subfigure}{0.19\columnwidth}
  \includegraphics[width=\linewidth]{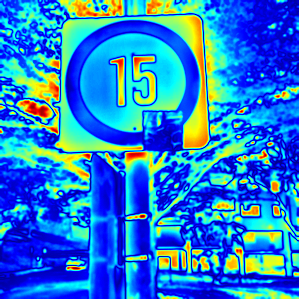}
  
\end{subfigure}
\begin{subfigure}{0.19\columnwidth}
  \includegraphics[width=\linewidth]{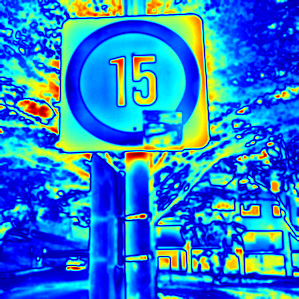}
  
\end{subfigure}
\begin{subfigure}{0.19\columnwidth}
\includegraphics[width=\linewidth]{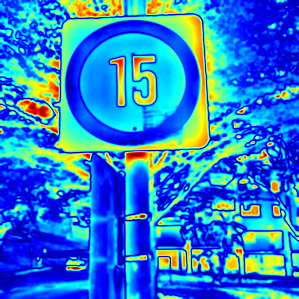}

\end{subfigure}

\caption{Adversarial patches and their saliency maps representing human attention. From the left to right, there are the original road sign, and adversarial patchs generated by GP (short for Google Patch)~\cite{Brown2017AdversarialP}, \re{LaVAN}~\cite{karmon2018lavan}, \re{PS-GAN}~\cite{Liu2019PerceptualSensitiveGF} and IAP. As shown in the saliency maps, IAP causes the least modifications, and thus has the greatest chance to evade detection from human.}
\label{fig:motivation}
\end{figure}

\IEEEPARstart{R}{ecent} years have witnessed the increasing trend of Deep Neural Networks (DNN) applied in many domains, which potentially play an essential role in constructing smart cities~\cite{zanella2014internet, jin2014information}.
Deep learning has dramatically improved the performances of many computer vision tasks and boosted the application of deep learning-based visual systems.
For example, image/face recognition systems are widely deployed on mobile phones nowadays for video surveillance~\cite{owayjan2015face}, access control~\cite{bhowmik2011thermal} and online payment~\cite{aria2020secure}. 
Well-known technological companies, including Tesla, Google, and Uber, have taken further steps to accelerate the commercialization of autonomous cars~\cite{deng2020analysis}.
Tencent is exploring to harnesses computer vision and IoT devices for smart agriculture~\cite{patricio2018computer}.

However, despite the popularity of deep learning, it has been shown susceptible to adversarial examples, which are crafted by adding negligible perturbations~\cite{Szegedy2014IntriguingPO}. 
Such minor perturbations can lead DNN's prediction into either an intentionally chosen class (targeted attack) or classes that are different from the true label (untargeted attack).
Extensive researches have studied to generate strong adversarial examples for classification problem~\cite{MoosaviDezfooli2016DeepFoolAS, Carlini_2017,papernot2016limitations,su2019one,Liu_2019_ICCV}, and soon extended to other high level vision tasks like object detection~\cite{ijcai2019-134}, object tracking~\cite{yan2020cooling}, and semantic segmentation~\cite{arnab2018robustness}.
To make attacks applicable in physical world, attacks based on adversarial patches are proposed~\cite{DBLP:journals/corr/KurakinGB16,Brown2017AdversarialP,Wiyatno_2019_ICCV,DBLP:conf/icml/LiSK19}. 
Different from adversarial perturbations added to whole images,
adversarial patches break the $l_p$ norm limitations for adversarial perturbations.
Consequentially, existing adversarial patches~\cite{Brown2017AdversarialP,karmon2018lavan,Liu2019PerceptualSensitiveGF} are ended being noticeable to the human observer because of their exotic appearance (see Fig.~\ref{fig:motivation}). 
Note that if the attacks are easily detected, such attacks are not applicable in most situations, and the success rates of attacks will be relatively low. 
Therefore, adversaries need to consider adversarial patches that are inconspicuous or even invisible to humans.
Liu~\mbox{\textit{et al.}}~\cite{Liu2019PerceptualSensitiveGF} observed there are often scrawls and patches on traffic signs in the real world, and they first attempted to enhance the visual fidelity while trying to preserve the attack ability.
Their method, however, is not effective as assumed, which is quite evident as shown in Fig.~\ref{fig:motivation}.
In addition, existing methods~\cite{Brown2017AdversarialP,karmon2018lavan,Liu2019PerceptualSensitiveGF} require a large amount of qualified data (e.g. ImageNet) for training, which is computationally expensive and time-consuming.
Apart from the high demand for computation resources, data scarcity is becoming an increasingly troublesome problem.
Sensitive data under privacy protection are often out of reach in many situations for training adversaries.
The above two problems make existing adversarial patch approaches inapplicable for practice.

Towards bridging the research gaps mentioned above, we propose a GAN-based approach in this work to generate Inconspicuous Adversarial Patches (IAP) with only one single image as the training data.
Our approach captures the most sensitive area of the victim images and generates adversarial patches with well-crafted objective functions.
The goals of IAP are (1) crafting adversarial patches with limited data and (2) evading human detection while keeping attacks successful.
Through experiments in digital settings, adversarial patches generated by our approach achieved high attack success rates.
Saliency detection and human evaluation show that our adversarial patches are less noticeable, and this approach highly reduces the risk of being detected.
In the last, we demonstrate the attack ability of IAP in physical scenarios to inspire future directions on developing real-world patch attacks.

In summary, the key contributions of this paper are as follows:
\begin{itemize}
    \item We are pioneering to study generating inconspicuous adversarial patches to evade detection to the best of our knowledge. We propose a novel approach IAP generates such adversarial patches from one image while preserving strong attack ability.
  \item We show the strong attack abilities of adversarial patches generated by IAP against standardly trained and adversarially trained models in different architectures and the application in the physical world through extensive experiments.
  \item We evaluate the detection risks of different adversarial patches by quantitative and qualitative analysis. It is demonstrated that adversarial patches generated by IAP, compared to existing adversarial patches, are consistent with backgrounds and able to evade detection by humans, which makes adversarial patches more dangerous.
\end{itemize}

In this paper, we extend our previous two-page abstract paper~\cite{Luo_Bai_Zhao_2021} with several notable improvements.
First, compared to~\cite{Luo_Bai_Zhao_2021}, we formalize the vulnerability map generation for patch selection. 
Second, we demonstrate the strong attack abilities of IAP on various adversarially trained models under different distances and give more samples generated by IAP.
Third, we carefully design the detection risk evaluation to show the inconspicuousness of adversarial patches generated by IAP, including saliency map visualization and human observation.
Four, we additionally incorporate the non-printability score loss in our objective function to make IAP applicable in the physical world.
Finally, we discuss the differences between inconspicuous adversarial patches and adversarial perturbations.

For the remaining part of this paper, we review the literature of adversarial attacks and defenses in Section~\ref{sec:Related Work}.
Then we introduce the problem definition and propose our approach in Section~\ref{sec:Our Approach}. 
Extensive experimental results are depicted in Section~\ref{sec:Experiment Results} and we give our conclusion in Section~\ref{sec:Conclusion}.

\section{Related Work}\label{sec:Related Work}
In this section, we review prior works of adversarial attacks and adversarial example detection.
\subsection{Adversarial Attacks}
Adversarial examples were first discovered \cite{Szegedy2014IntriguingPO}.
Szegedy~\mbox{\textit{et al.}}~\cite{Szegedy2014IntriguingPO} added some imperceptible noises on the clean image, and misled well-trained classification models successfully.
This greatly attracted researchers' attention and extensive attacks are developed since then.
According to the restrictions when crafting adversarial examples, adversarial examples can be roughly categorized into three types: perturbation-based, patch-based, and unrestricted adversarial examples.

\subsubsection{Adversarial Perturbation.} 
Intrigued by~\cite{Szegedy2014IntriguingPO}, many perturbation-based attacks are proposed in recent years.
\cite{43405} proposed Fast Gradient Sign Method (FGSM) to generate adversarial examples fast and effectively.
Inspired by single-step FGSM, Authors of~\cite{Tramr2018EnsembleAT,Kurakin2017AdversarialEI} enhanced FGSM with a randomization step or multiple gradient steps.
More representative attacks like \cite{papernot2016limitations,MoosaviDezfooli2016DeepFoolAS,Carlini_2017,madry2018towards,9165820} explore possibilities of attacks in different ways.
Papernot~\mbox{\textit{et al.}}~\cite{papernot2016limitations} proposed to generate adversarial examples guided by a saliency map, while Moosavi-Dezfooli~\mbox{\textit{et al.}}~\cite{MoosaviDezfooli2016DeepFoolAS} proposed Deepfool based on the decision boundary.
Carlini and Wagner~\cite{Carlini_2017} formulated the attack as an optimization problem and proposed a novel regularization term to enhance the attacks.
Madry~\mbox{\textit{et al.}}~\cite{madry2018towards} employed Projected Gradient Descent (PGD) to generate adversarial examples, which is currently known as the strongest first-order attack.
But these methods share common restrictions on the norm of perturbations, where a noise budget $\epsilon$ is given.
The goal of such attacks is to find adversarial examples with small $\epsilon$ so that the perturbations are ensured to be imperceptible to human eyes.

\subsubsection{Adversarial Patch.} 
Adversarial patches are inspired by the pixel attack~\cite{su2019one} that changes only one pixel to an arbitrary value.
However, the success rate of pixel attacks is relatively low because such changes are minor on high resolution images.
Patch-based attacks break the restriction in pixel attack that only one pixel is replaced, which makes the attack more efficient and applicable in physical world.
Adversarial patches, were first generated by~\cite{Brown2017AdversarialP} in 2017.
By masking relatively small patches on the image, such attack causes classifiers to ignore other scenery semantics and report a false prediction. 
Eykholt~\mbox{\textit{et al.}}~\cite{inproceedings2018patch} generates robust patches under different physical conditions to fool the classification of real-world road sign. 
Some authors also investigated adversarial patches for attacking object detection models tested in digital settings \cite{Liu2019DPATCHAA} and physical settings \cite{Thys_2019,Lee2019OnPA}. 

Though existing adversarial patches have great attack ability, they are highly conspicuous.
If there is any detection, such patches will be spotted, leading to failures of the attacks.
To make adversarial patches bypass potential detection, \cite{Liu2019PerceptualSensitiveGF} generates more realistic patches with Generative Adversarial Nets (GAN)~\cite{10.5555/2969033.2969125}.
In~\cite{jia2020adv}, the malicious area is disguised as watermarks to evade detection. 
Their approaches assume that people's understanding of the image content is not affected by such semi-transparent perturbations and hence people's consciousness will not be aroused.

\subsubsection{Unrestricted Adversarial Examples}
When generating the above discussed adversarial examples, there are always restrictions like the perturbations' norm or the patch size. The reason that such restrictions exist is that attacks naturally have to be undetectable or difficult to detect. Unrestricted adversarial examples occur along with the quick development of GAN, since identifying images synthesized GAN from natural images in a glance becomes impossible for humans. Unrestricted adversarial examples are first discovered by~\cite{NIPS2018_8052}. They employed GAN to approximate the distribution of training data, then searched for adversarial examples within the distribution. 
Qiu~\mbox{\textit{et al.}}~\cite{qiu2019semanticadv} directly manipulate latent vectors of images which contains semantic information, and inputs the latent vector to GAN so that adversarial examples are disguised as natural images.

\subsection{Adversarial Example Detection}
In practice, adversarial example detection is one of effective defenses on adversarial examples. 
Lu~\mbox{\textit{et al.}}~\cite{lu2017safetynet} assumed adversarial examples produce different patterns of activation functions, proposed to append a SVM classifier to detect adversarial perturbations.
Similarly, authors of~\cite{metzen2017detecting,meng2017magnet} attached subnetworks which are trained for classification of benign and malicious data.
Statistics of convolutions in CNN-based networks are used for detection in~\cite{li2017adversarial}.
Feature squeezing is also an effective approach to detect adversarial perturbations~\cite{xu2017feature}.

However, to our best knowledge, the current research still lacks comprehensive studies on automatic detection of adversarial patches trained on large-scale datasets. One possible reason is that existing adversarial patches preferably emphasize on the attack ability, so they cannot even escape from human observation.
Our paper would be a pioneering work in which adversarial patches are imperceptible to some extent while preserving strong attack ability. 
We hope our work could shed some light on benchmarks of inconspicuous patch-based attacks in the future, and inspire researchers to develop defense algorithms robust to adversarial patches.

\section{Our Approach}\label{sec:Our Approach}
In this section, we introduce the problem definition and elaborate the framework and formulation of our proposed patch-based attack named IAP.
\subsection{Problem Definition}
Patch-based attack is created by completely replacing a part of original image.
Given a victim image $x$ of size $(H, W)$ whose label is $y$, we aim to generate an adversarial patch $p$ of size $(h,w)$, which can result in prediction errors of well-trained target models $f:[-1,1]^{d} \rightarrow \mathbb{R}^{C}$, where $d$ is the input dimension, $C$ is the number of classes. 
We attach $p$ with a location mask $m\in\{0,1\}_{(H \times W)}$ onto the victim image.
The new image is given by
 \begin{equation}
 \begin{aligned}
x' = m\otimes p + (1-m)\otimes x,
 \label{advp masking definition}
 \end{aligned}
 \end{equation}
where $\otimes$ is the element-wise multiplication for matrices, and $p$ is padding to same size of $x$.
For the ease of reading, we replace the masking process described in Equation~\ref{advp masking definition} with an abstract notion $\oplus$ in the following sections:
 \begin{equation}
 \begin{aligned}
& x' = p \oplus x.
 \label{advp definition simplified}
 \end{aligned}
 \end{equation}
We call $p$ an adversarial patch if it satisfies
 \begin{equation}
 \begin{aligned}
y \neq \underset{c=1,2, \ldots, C}{\arg \max } f(x').
 \label{advp definition}
 \end{aligned}
 \end{equation}

With such an adversarial patch, the output of $f$ on $x'$ is expressed as $f(x')$, which differs from $f(x)$.

\subsection{Vulnerability Map Generation}
Since convolutional neural networks are known to be perceptually sensitive to certain semantics (points of interest) that contribute to the correct prediction, we use the gradient-based explainability mechanism \cite{DBLP:journals/corr/SelvarajuDVCPB16} to enhance the attacking capability of our approach. The mechanism highlights the regions of input that are most important to predictions of models. Specifically, the mechanism computes the gradient of the prediction  score $y$ for a class $c$ w.r.t. feature maps $A$. Such feature maps ($k$ pieces in total) are extracted from convolutional layers of the deep model. These gradients are globally average pooled to output a weight $\alpha_{k}^c$ (the feature importance): 
\begin{equation}
\begin{aligned}
\alpha_{k}^c = \frac{1}{Z}\sum_{i}\sum_{j}\frac{\partial y^{c}}{\partial A_{ij}^k}.
\end{aligned}
\end{equation}
Then the sensitivity heatmap $M$ for the class $c$ is generated by calculating the ReLU output of the weighted sum of those feature maps, which is expressed as
\begin{equation}
\begin{aligned}
M^c = ReLU(\sum_{k} \alpha_{k}^c A^k).
\end{aligned}
\end{equation}

\begin{figure}[t]
    \centering
    \includegraphics[width=\columnwidth]{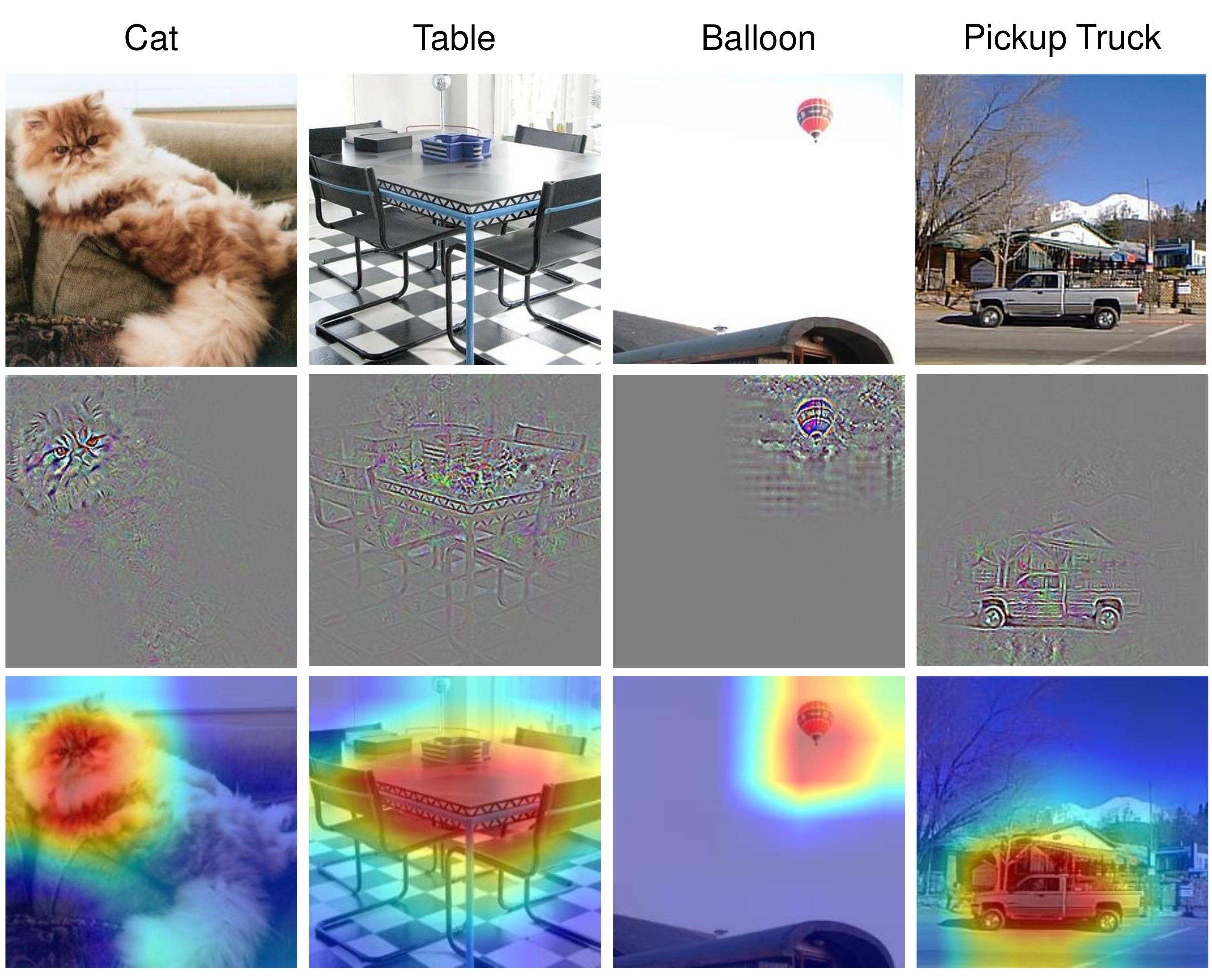}
    \caption{Generated examples of gradient-based explainability mechanism for a typical CNN-based model ResNet50. The first row shows original images, the second row shows the computed gradients and the third row are the heatmaps applied on original images.}
    \label{gradcam}
\end{figure}

Based on the sensitivity heatmap $M$, we choose the area with the highest importance as the victim area for patch attack.
We show some examples of this mechanism in Fig.~\ref{gradcam}. 
For implementation, we refer to Grad-CAM's specification and details in \cite{DBLP:journals/corr/SelvarajuDVCPB16}.

\begin{figure*}[t]
    \centering
    \includegraphics[width=\textwidth]{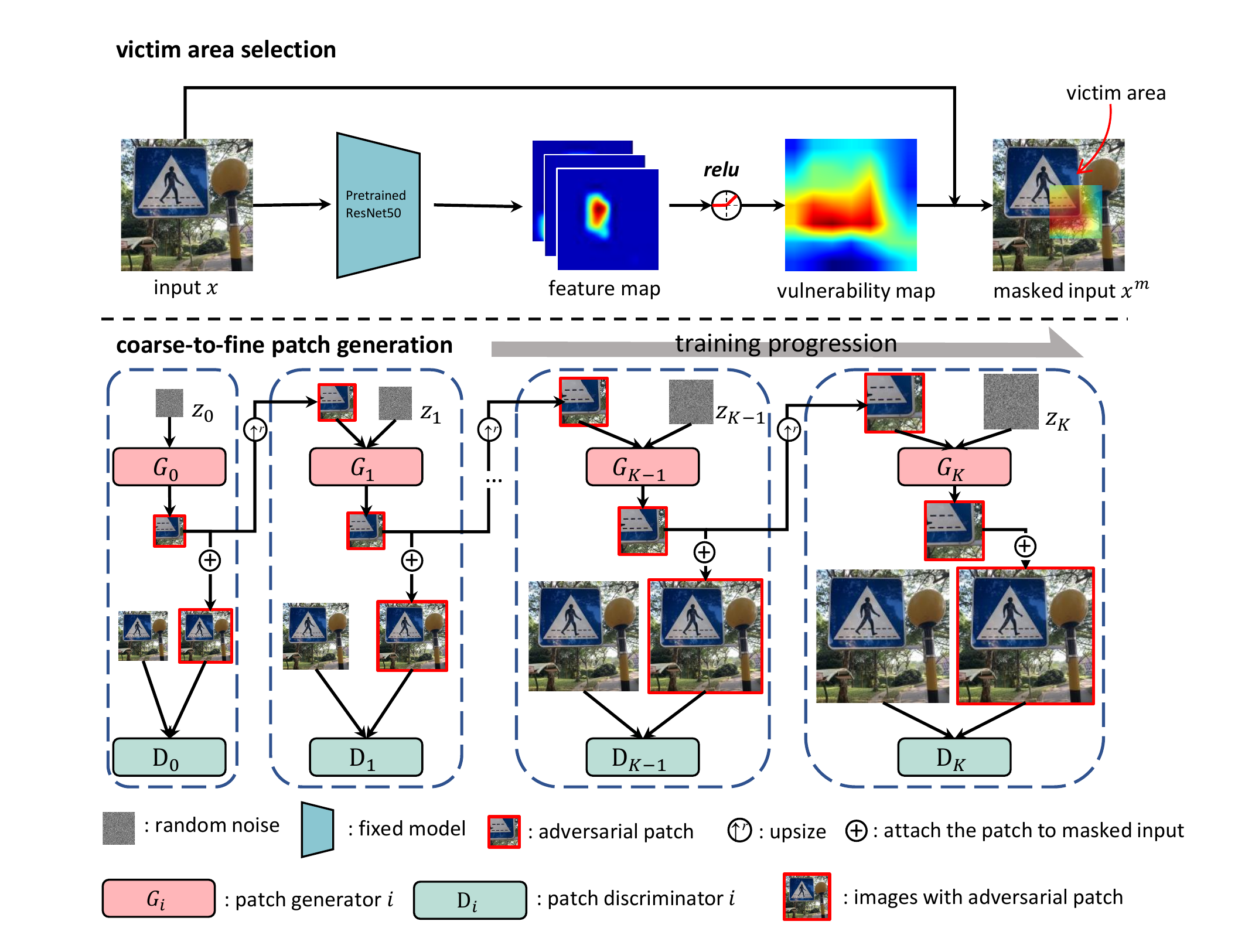}
    \caption{The Overall Structure of Our Approach. First, given an input image, we select the victim area based on the attention map of the input for adversarial patch location. Then we perform a coarse-to-fine generation process to generate adversarial patches. More details can be found in Section~\ref{sec:iap framework}.}
    \label{approach}
\end{figure*}

\subsection{IAP Framework}\label{sec:iap framework}
The overview of our framework is illustrated in Fig.~\ref{approach}.
Given an image $x$, we first generate the vulnerability map with pre-trained models, which captures the sensitivity of models' output with respect to the input image. 
In white-box settings, we use the target models, while in black-box settings, we select one of the available pre-trained models arbitrarily.
The areas in red are better candidates recommended to attack than areas in blue.
Here we take the attack ability and visual fidelity into account and adopt a coarse-to-fine process using GAN with different scales for inconspicuous adversarial patch generation.
The generators in each scale (except the first scale) generate adversarial patches from noises and the patches generated from the last scale. The discriminators classify the downsampled input images and images with patches (except the last scale).
Different from existing methods, our approach takes only one single image as training data.

\subsubsection{Multi-Scale Patch Generation}
In our approach, we deploy a series of generator-discriminator pairs $\{(G_0, D_0),\dots,(G_K,D_K)\}$, where $K$ is the total number of scales in the structure.
As shown in Fig.~\ref{approach}, these generator-discriminator pairs are trained on an image pyramid of patches and downsampled images.
To be specific, the image pyramid is expressed as $\left\{(p_0,x_0)\dots(p_K,x_K)\right\}$, where $p_i$ and $x_i$ are downsampled version of $p$ and $x$ with a factor $r^{K-i}$ $(0<r<1)$.
On every scale, we execute adversarial training for the generator and the discriminator.
The generator $G_i$ is expected to produce realistic patches, and the discriminator attempts to distinguish the images with patches $x'_i = p_i \oplus x_i$ from the original images $x_i$.
To make the generated patches consistent with the original images, we attach the patches to the images in every scale. 
During training, the generators will learn the information from original images progressively.

As we can see from Fig.~\ref{approach}, the generation of patches happens in a sequential way, which starts at the coarsest scale and ends at the finest scale.
Generators take the output from the previous coarser scales and noises as the new input except the first generator.
Specifically, the generated patch from coarser scales is upsampled and serves as a prior for the generators in finer scales.
Random noises $z$ are used for generation diversity.
Finally, our generator at the finest scale $G_K$ will output adversarial patches $\tilde{p}$ of size $(h,w)$ on a prospective location $L$ of the victim image $x$.

\subsubsection{Objective Functions}
As presented above, each generator is coupled with a Markovian discriminator~\cite{li2016precomputed}, and the training structure of different scales are the same.
We take the $i_{th}$ scale to elaborate the training objectives.

We denote the output of $G_{i-1}$ as $\tilde{p}_{i-1}$, then the input for $G_i$ is 
\begin{equation}\tilde{p}_{i}=G_{i}\left(z_{i},\left(\bar{p}_{i-1}\right) \uparrow^{r}\right), \end{equation}
where $\bar{p}_{i-1} \uparrow^{r}$ is the upsampled version of $\tilde{p}_{i-1}$.

The adversarial loss~\cite{10.5555/2969033.2969125} for training generators and discriminators can be written as
\begin{equation}
\begin{aligned}
\mathcal{L}_{\mathrm{GAN}} = &\mathbb{E}_{p_i \sim x} \log \mathcal{D}(\tilde{p}_{i}, x_i)+ \\ 
 & \mathbb{E}_{z_i \sim \mathcal{P}_z} \log (1-\mathcal{D}(\mathcal{G}(z_i, \bar{p}_{i-1} \uparrow^{r}), x_i)),
\end{aligned}
\end{equation}
where $\mathcal{P}_z$ is a prior for noises.

The loss for fooling target model $f$ in untargeted attacks is
\begin{equation}
\mathcal{L}_{\mathrm{adv}}^{f}=\mathbb{E}_{x} \ell_{f}(x \oplus \bar{p}_{i} \uparrow^{r}, y),
\end{equation}
where $\ell_{f}$ denotes the loss function used in the training of $f$, and $y$ is the true class of $x$.
Note that it is feasible to replace $y$ with labels of other classes to perform targeted attacks.

To stabilize the training of GAN, we add the reconstruction loss
\begin{equation}
\mathcal{L}_{\mathrm{rec}}=\left\|G_{i}\left(z_i, \bar{p}_{i-1} \uparrow^{r}\right)-p_{i}\right\|^{2}.
\end{equation}

We also add a total variation loss
 \begin{equation}
 \begin{aligned}
 \mathcal{L}_{\mathrm{tv}} = \sum_{a=0}^h \sum_{b=0}^w (\left|\tilde{p}_i^{(a+1,b)}-\tilde{p}_i^{(a,b)}\right|+\left|\tilde{p}_i^{(a,b+1)}-\tilde{p}_i^{(a,b)}\right|)
 \end{aligned}
 \end{equation}
as a regularization term to ensure that the texture of generated patches is smooth enough.

Overall, the objective function in $i_{th}$ scale can be expressed as
\begin{equation}\mathcal{L}=\mathcal{L}_{\text {adv }}^{f}+\alpha \mathcal{L}_{\mathrm{GAN}}+\beta \mathcal{L}_{\mathrm{rec}}+\gamma \mathcal{L}_{\mathrm{tv}},
\label{eqn:overall loss}
\end{equation}
where $\alpha$, $\beta$ and $\gamma$ are to balance the relative importance of each loss.
Then we train our generator and discriminator by solving the min-max game, which is expressed as 
\begin{equation}
\argmin_{G_{i}} \max _{D_{i}} \mathcal{L}\left(G_{i}, D_{i}\right).
\end{equation}

Overall, our generative models are trained sequentially, starting from the coarsest scale.
For each scale, the parameters of GAN will be fixed once training is finished.
Then the training repeats in the next scale until the finest scale.
The training details are summarized in Algorithm~\ref{alg::pipeline}.

\begin{algorithm}[tbp]
\SetAlgoLined
\KwInput{input image $(x, y)$, target model $f$}
\KwOutput{generators and discriminators $(\theta_{G_i}, \theta_{D_i}), \text{where \ } i \in \{1\dots K\}$}
Generate image pyramid$\left\{(p_0,x_0)\dots(p_K,x_K)\right\}$ \;
\For{each scale $i \in \{1\dots K\} $}{
    \For{each training iteration}{
        \For{$S_D$ steps}{
            sample $z_i \sim \mathcal{P}_{z}$ \;
            \eIf{i = 1}{
                $\tilde{p}_i \leftarrow G_i(z_i)$ \;
            }{
                $\tilde{p}_i \leftarrow G_i(z_i, \bar{p}_{i-1} \uparrow^{r})$ \;
            }
            $x'_{i} \leftarrow \tilde{p}_i \oplus x_i$ \;
            $\theta_{D_i} \leftarrow \theta_{D_i} + r\nabla\left(\mathcal{L}_{GAN}(x'_{i}, x_{i})\right)$ \;
        }
        \For{$S_G$ steps}{
            sample $z_i \sim \mathcal{P}_{z}$ \;
            \eIf{i = 1}{
                $\tilde{p}_i \leftarrow G_i(z_i)$ \;
            }{
                $\tilde{p}_i \leftarrow G_i(z_i, \bar{p}_{i-1} \uparrow^{r})$ \;
            }
            $\theta_{G_i} \leftarrow \theta_{G_i} + r\nabla\left(\mathcal{L}(\tilde{p}_i, p_i, x_{i})\right)$ \;
        }
    }
}

 \caption{IAP Generation}
 \label{alg::pipeline}  
\end{algorithm}

\begin{table*}[h]
\caption{White-box and Black attack success rate for every category's 10 models. The white-box is conducted on InceptionV3 and black-box is then conducted on GoogleNet, MNASNet (multiplier of 1.0), MobileNetV2, robust InceptionV3, and robust MobileNetV2. The statistics indicate that our approach has strong attacking ability in white-box and black-box settings.}
\label{success rate}
\normalsize
\centering
\begin{tabular}{lcccccc}
\toprule 
 & White-box attack & \multicolumn{5}{c}{Black-box attack} \\ \cmidrule(r){2-2} \cmidrule(r){3-7}
Class Index/Name       & InceptionV3 & GoogleNet & MNASNet & MobileNetV2 & R-InceptionV3 & R-MobileNet\\ \midrule
\textbf{283 Persian Cat}      & 100.00\%             & 99.22\%            & 85.62\%          & 90.66\%            & 68.10\%          & 80.33\%                     \\ %
\textbf{340 Zebra}            & 98.53\%              & 99.36\%            & 85.58\%          & 90.38\%            & 73.62\%          & 74.40\%                      \\ %
\textbf{417 Balloon}          & 99.52\%              & 99.19\%            & 79.68\%          & 90.19\%            & 67.91\%          & 82.70\%                     \\ %
\textbf{527 Desktop Computer} & 99.72\%              & 99.20\%            & 82.23\%          & 90.71\%            & 23.09\%          & 77.86\%                     \\ %
\textbf{532 Table}            & 99.95\%              & 99.19\%            & 86.13\%          & 90.09\%            & 69.41\%          & 87.09\%                     \\ %
\textbf{604 Hourglass}        & 99.99\%              & 99.20\%            & 82.12\%          & 90.59\%            & 60.42\%          & 80.39\%                      \\ %
\textbf{717 Pickup Truck}     & 99.97\%              & 99.31\%            & 85.34\%          & 91.06\%            & 48.04\%          & 75.65\%                     \\ %
\textbf{919 Street Sign}      & 98.30\%              & 99.21\%            & 82.27\%          & 90.10\%            & 79.99\%          & 84.92\%                      \\ %
\textbf{964 Potpie}           & 99.88\%              & 99.30\%            & 83.81\%          & 90.85\%            & 32.39\%          & 80.01\%                      \\ %
\textbf{975 Lakeside}         & 99.91\%              & 99.30\%            & 84.00\%          & 89.96\%            & 78.86\%          & 71.85\%                      \\ \midrule
\textbf{Average}              & 99.58\%              & 99.25\%            & 83.68\%          & 90.46\%            & 60.18\%          & 79.52\%                      \\ \bottomrule
\end{tabular}%
\end{table*}

\section{Experiment Results}\label{sec:Experiment Results}
This section conducted a series of experiments to evaluate the attack ability in both white-box and black-box settings.
To assess the inconspicuousness of the generated patches, we make comparisons of saliency detection and human evaluation on different types of adversarial patches. We also attempt physical attacks to evaluate our approach in the real-world scenarios.
\subsection{Implementation Details}
We adopt similar architectures of generators used in~\cite{Isola_2017,zhu2017unpaired} and adopt a fully convolutional network with 5 conv-blocks as discriminators~\cite{Shaham_2019}.
In adversarial training, we use WGAN-GP loss in~\cite{gulrajani2017improved} because it provides higher stability and generation quality.
Since C\&W loss~\cite{Carlini_2017} is proved to be effective to generate strong adversarial exmaples, we adopt it as the attack loss $\mathcal{L}^{f}_{adv}$.
Our approach is implemented on a workstation with four GPUs of NVIDIA GeForce RTX 2080 Ti. 
The optimizer for all generators and discriminators is Adam with a learning rate 0.0005.

\re{For deployment, given the victim image, we train our generators to generate adversarial patches with the help of saliency map. 
Then we attach the adversarial patches to the victim image on the victim area.}

\subsection{White-box and Black-box Attack}
In this section, our data are randomly sampled from ImageNet~\cite{imagenet_cvpr09}.
Due to resource limitation and the nature of IAP, it is infeasible for us to use the whole dataset in our experiments.
Therefore, we choose 10 classes from ImageNet, and sample 10 images in each class (see details in Table~\ref{success rate}).
In total, 100 images are used as our evaluation set.
To assess the attack capability of adversarial patches generated by our approach, we conduct experiments in white-box settings and black-box settings respectively.
\re{The success rate in our paper is defined as the number of samples misclassified over the total number of samples.}

\begin{figure*}[h]
    \centering
    \includegraphics[width=\textwidth]{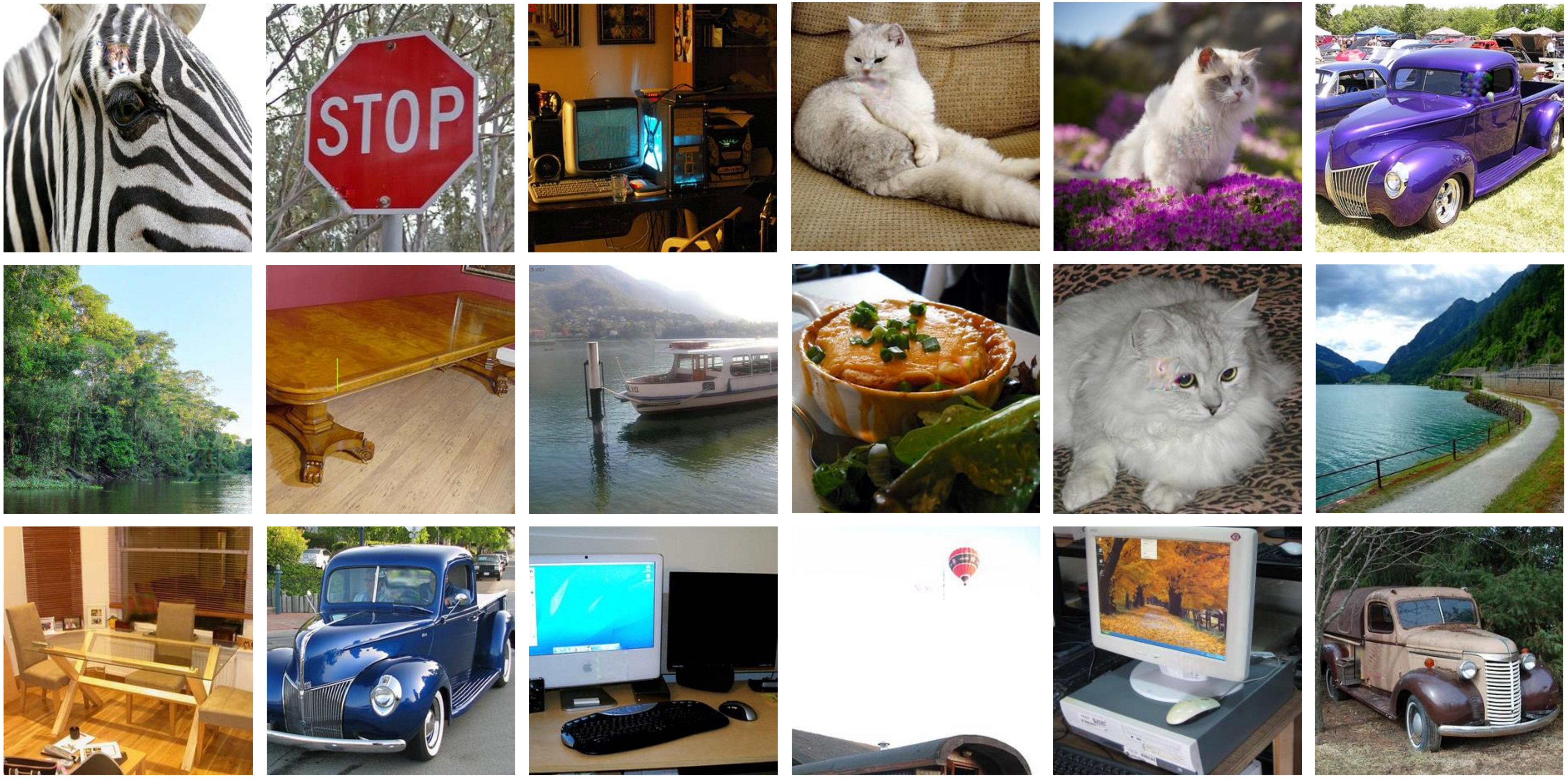}
    \caption{Examples of adversarial patches generated by our approach. At the first glance, most of the patches are inconspicuous to human observers.}
    \label{sample demo}
\end{figure*}

\subsubsection{White-box Attack} 
We use Inception V3~\cite{DBLP:journals/corr/SzegedyVISW15} as the target model for white-box attacks.
The size of generated patches is fixed as (3, 40 ,40), which cover less than 2\% area of the input images (the input size of Inception V3 is (3, 299, 299)).
Since our approach runs on one single image, we trained 100 models with all images sampled above.
Within each trained model, we generated 1,000 adversarial patches for evaluation.

The categorical white-box success rates are shown in the first column of Table~\ref{success rate}.
We can see that the attacks are of exceptionally high success rates with a 99.58\% on average.
We compare our method with three SOTA methods: Google AP, \re{LaVAN}, and PSGAN. 
IAP shows comparable or even stronger attack abilities, and the results are summarized in Table~\ref{baseline performance}. 
A few randomly selected examples generated by our method are illustrated in Fig.~\ref{sample demo}.

\subsubsection{Transferability for Black-box Attack}
In black-box settings, the adversaries are not allowed to access the target models.
To perform attacks, they can only produce adversarial patches with known models, then use these patches for attacks.
Thus we use the 100,000 adversarial patches generated in white-box settings, where the target model is Inception V3.
There are four different target black-box models for evaluation, namely GoogLeNet \cite{Szegedy_2015}, MNASNet \cite{Tan_2019} with a depth multiplier as 1.0, and MobileNetV2 \cite{Sandler_2018}.
The average success rates of our patches on these three models are 99.25\%, 83.68\%and 90.46\%, respectively (see Table~\ref{success rate}).

\begin{table}[t]
\caption{The attack performance comparison among several approaches.}
\label{baseline performance}
\centering
\normalsize
\begin{tabular}{ccccc}
\toprule
             & InceptionV3 & MNASNet & GoogleNet &\\ \midrule
Google Patch & 56.94\%     & 72.59\% & 97.20\%   &  \\
LaVAN        & 48.33\%     & 83.30\% & 99.00\%   &  \\
PSGAN        & 88.01\%     & 83.54\% & \textbf{99.50\%}   &  \\
IAP          & \textbf{99.58\%}     & \textbf{83.68\%} & 99.25\%   &  \\
\bottomrule
\end{tabular}

\end{table}

In addition, we test our adversarial patches on adversarially-trained models to demonstrate the strong attack ability of IAP. 
We select two adversarially trained models on ImageNet: Robust InceptionV3\footnote{Downloaded from  https://github.com/fastai/timmdocs} and Robust MobileNetV2\footnote{Downloaded from https://github.com/microsoft/robust-models-transfer}.
Adversarial training of Robust InceptionV3 is under $l_{\infty}$ distance with $\epsilon = 16/ 255$, and Robust MobileNetV2 is trained under $l_{2}$ distance with $\epsilon = 3$.
In black-box settings, the average success rates IAP achieves are $60.18\%$ and $79.52\%$, which we believe is quite good.
Details are shown in Table~\ref{success rate}.

\subsection{Detection Risk Evaluation}
In this section, we evaluate the risks of adversarial patches prone to human detection from two aspects.
Qualitatively, we visualize saliency maps to show the simulated human-attention focus area.
Quantitatively, we conduct human evaluations on different types of adversarial patches.

\subsubsection{Visual Saliency Detection}
Saliency map~\cite{Montabone2010HumanDU} is developed to simulate human's attentions based on a biologically-inspired attention system.
The simulated human's attention is highlighted in the target image to indicate the observer's interest when observing the images. 
With saliency maps, we can easily approximate what people are focusing on when looking at the images.

\begin{figure*}[t]
\centering

\begin{subfigure}{0.19\columnwidth}
\caption*{Original}
  \includegraphics[width=\linewidth]{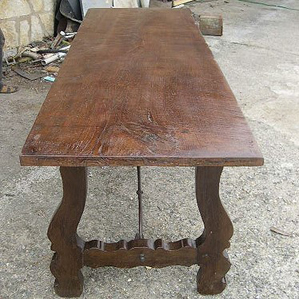}
\end{subfigure}
\begin{subfigure}{0.19\columnwidth}
\caption*{GP}
  \includegraphics[width=\linewidth]{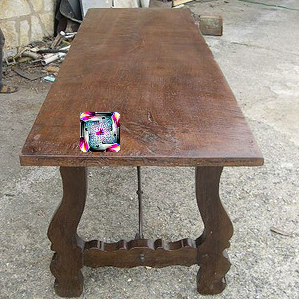}
\end{subfigure}
\begin{subfigure}{0.19\columnwidth}
\caption*{LaVAN}
  \includegraphics[width=\linewidth]{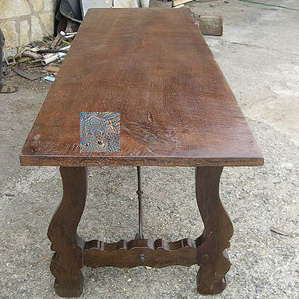}
\end{subfigure}
\begin{subfigure}{0.19\columnwidth}
\caption*{\re{PS-GAN}}
  \includegraphics[width=\linewidth]{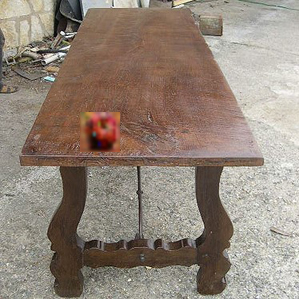}
\end{subfigure}
\begin{subfigure}{0.19\columnwidth}
\caption*{IAP (Ours)}
\includegraphics[width=\linewidth]{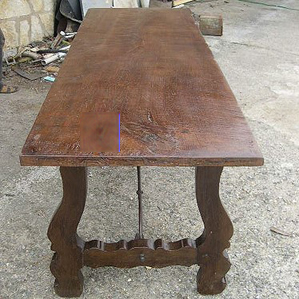}
\end{subfigure}
\begin{subfigure}{0.19\columnwidth}
\caption*{Original}
  \includegraphics[width=\linewidth]{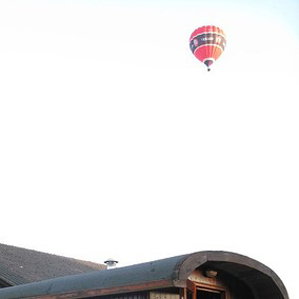}
\end{subfigure}
\begin{subfigure}{0.19\columnwidth}
\caption*{GP}
  \includegraphics[width=\linewidth]{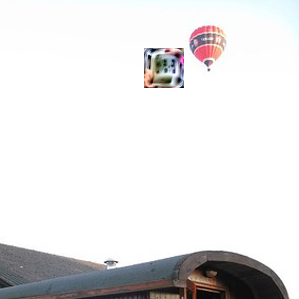}
\end{subfigure}
\begin{subfigure}{0.19\columnwidth}
\caption*{LaVAN}
  \includegraphics[width=\linewidth]{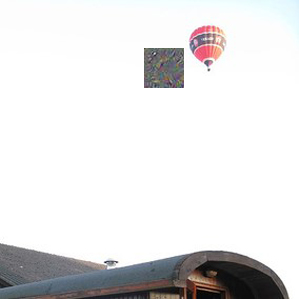}
\end{subfigure}
\begin{subfigure}{0.19\columnwidth}
\caption*{\re{PS-GAN}}
  \includegraphics[width=\linewidth]{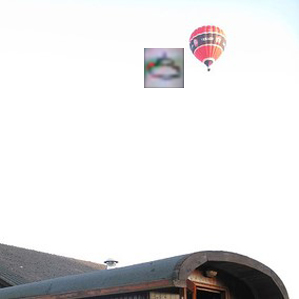}
\end{subfigure}
\begin{subfigure}{0.19\columnwidth}
\caption*{IAP (Ours)}
\includegraphics[width=\linewidth]{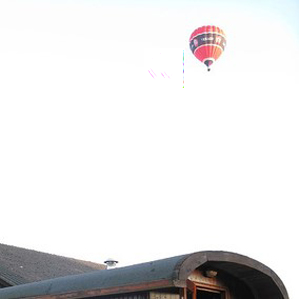}
\end{subfigure}

\vspace{.4em}

\begin{subfigure}{0.19\columnwidth}
  \includegraphics[width=\linewidth]{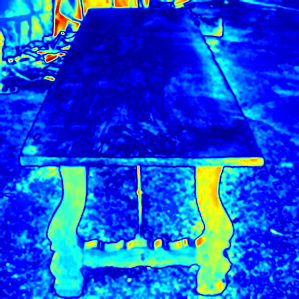}
\end{subfigure}
\begin{subfigure}{0.19\columnwidth}
  \includegraphics[width=\linewidth]{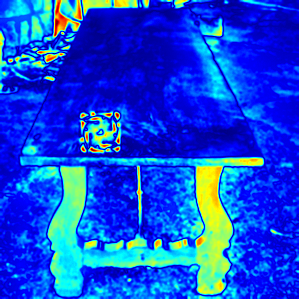}
  
\end{subfigure}
\begin{subfigure}{0.19\columnwidth}
  \includegraphics[width=\linewidth]{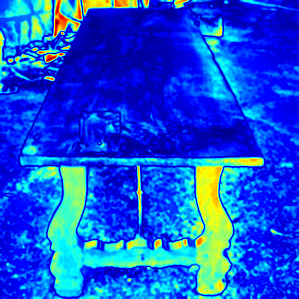}
  
\end{subfigure}
\begin{subfigure}{0.19\columnwidth}
  \includegraphics[width=\linewidth]{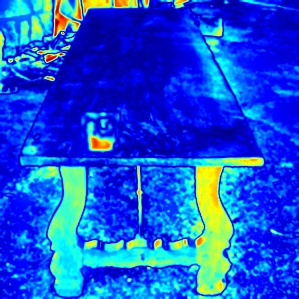}
  
\end{subfigure}
\begin{subfigure}{0.19\columnwidth}
\includegraphics[width=\linewidth]{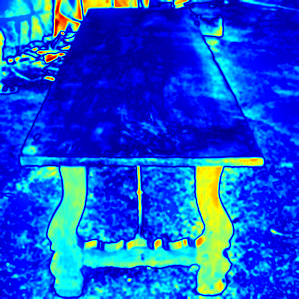}
\end{subfigure}
\begin{subfigure}{0.19\columnwidth}
  \includegraphics[width=\linewidth]{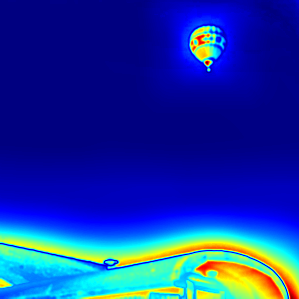}
\end{subfigure}
\begin{subfigure}{0.19\columnwidth}
  \includegraphics[width=\linewidth]{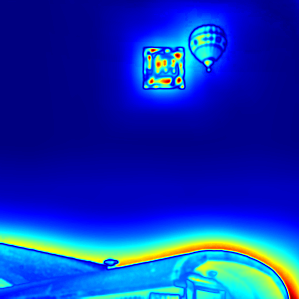}
  
\end{subfigure}
\begin{subfigure}{0.19\columnwidth}
  \includegraphics[width=\linewidth]{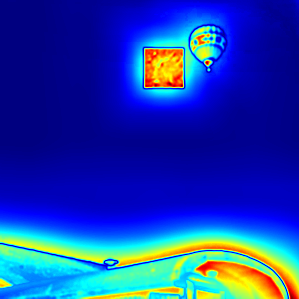}
  
\end{subfigure}
\begin{subfigure}{0.19\columnwidth}
  \includegraphics[width=\linewidth]{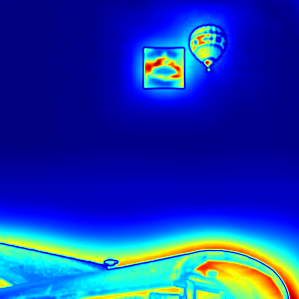}
  
\end{subfigure}
\begin{subfigure}{0.19\columnwidth}
\includegraphics[width=\linewidth]{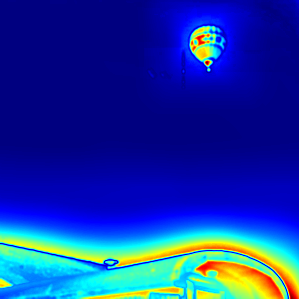}
\end{subfigure}

\vspace{.4em}%

\begin{subfigure}{0.19\columnwidth}

  \includegraphics[width=\linewidth]{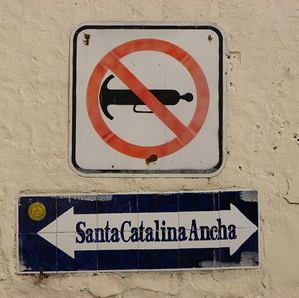}
\end{subfigure}
\begin{subfigure}{0.19\columnwidth}

  \includegraphics[width=\linewidth]{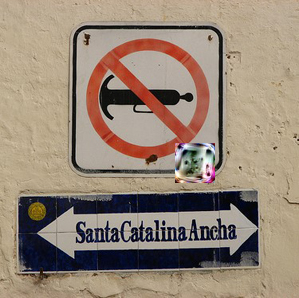}
\end{subfigure}
\begin{subfigure}{0.19\columnwidth}

  \includegraphics[width=\linewidth]{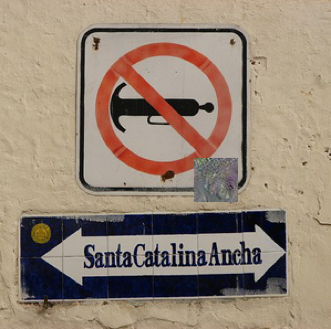}
\end{subfigure}
\begin{subfigure}{0.19\columnwidth}

  \includegraphics[width=\linewidth]{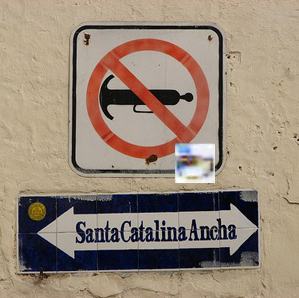}
\end{subfigure}
\begin{subfigure}{0.19\columnwidth}

\includegraphics[width=\linewidth]{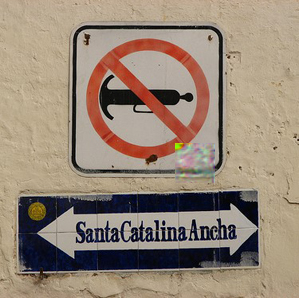}
\end{subfigure}
\begin{subfigure}{0.19\columnwidth}

  \includegraphics[width=\linewidth]{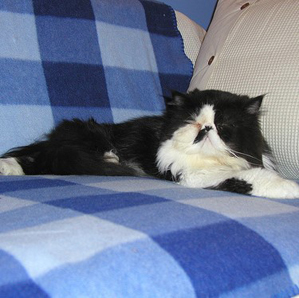}
\end{subfigure}
\begin{subfigure}{0.19\columnwidth}

  \includegraphics[width=\linewidth]{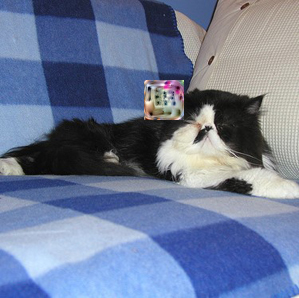}
\end{subfigure}
\begin{subfigure}{0.19\columnwidth}

  \includegraphics[width=\linewidth]{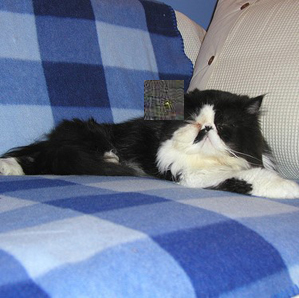}
\end{subfigure}
\begin{subfigure}{0.19\columnwidth}
 \includegraphics[width=\linewidth]{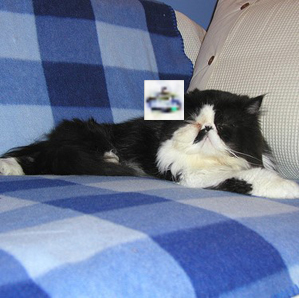}
\end{subfigure}
\begin{subfigure}{0.19\columnwidth}

\includegraphics[width=\linewidth]{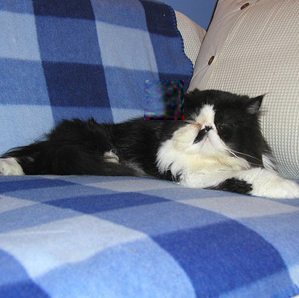}
\end{subfigure}

\vspace{.4em}

\begin{subfigure}{0.19\columnwidth}
  \includegraphics[width=\linewidth]{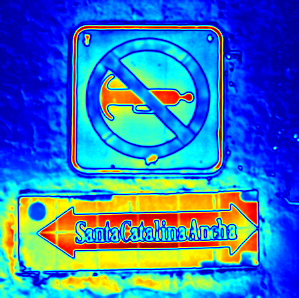}
\end{subfigure}
\begin{subfigure}{0.19\columnwidth}
  \includegraphics[width=\linewidth]{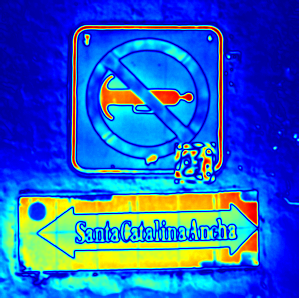}
  
\end{subfigure}
\begin{subfigure}{0.19\columnwidth}
  \includegraphics[width=\linewidth]{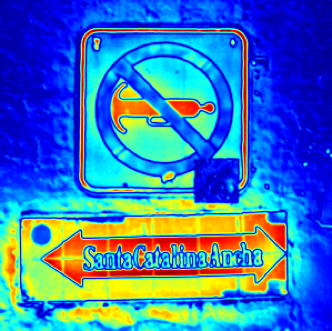}
  
\end{subfigure}
\begin{subfigure}{0.19\columnwidth}
  \includegraphics[width=\linewidth]{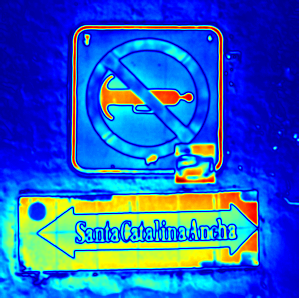}
  
\end{subfigure}
\begin{subfigure}{0.19\columnwidth}
\includegraphics[width=\linewidth]{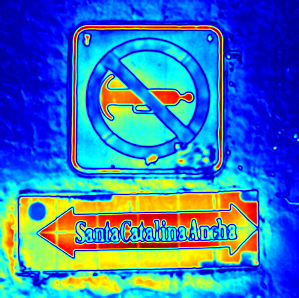}
\end{subfigure}
\begin{subfigure}{0.19\columnwidth}
  \includegraphics[width=\linewidth]{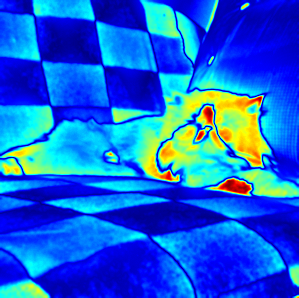}
\end{subfigure}
\begin{subfigure}{0.19\columnwidth}
  \includegraphics[width=\linewidth]{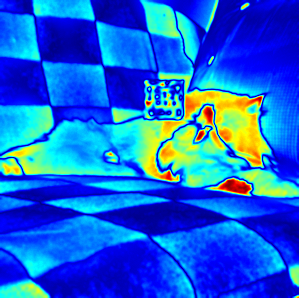}
  
\end{subfigure}
\begin{subfigure}{0.19\columnwidth}
  \includegraphics[width=\linewidth]{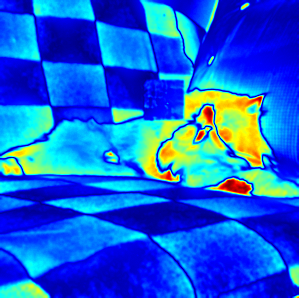}
  
\end{subfigure}
\begin{subfigure}{0.19\columnwidth}
  \includegraphics[width=\linewidth]{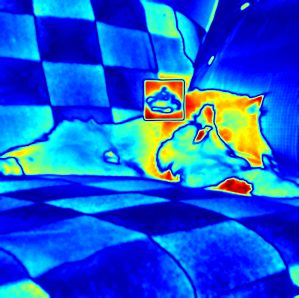}
  
\end{subfigure}
\begin{subfigure}{0.19\columnwidth}
\includegraphics[width=\linewidth]{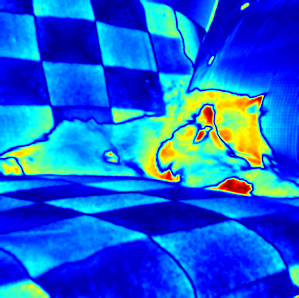}
\end{subfigure}

\caption{Examples of saliency detection for adversarial patches. We can see that our approach well reserves the visual consistency compared to other approaches and does not trigger the saliency detection significantly.}
\label{Examples of saliency detection}
\end{figure*}

We compete our synthetic patches with Google Patch \cite{Brown2017AdversarialP}, \re{PS-GAN} \cite{Liu2019PerceptualSensitiveGF}, and LaVAN \cite{karmon2018lavan} since these patch-based adversaries are targeted at image classifiers for digital images. 
We include original images as the reference.
Note that, all the patches are attached in the same location for the fairness of comparison.
A few examples and their saliency maps are illustrated in Fig.~\ref{Examples of saliency detection}.
We can see that patches generated by the other three approaches can be easily spotted, and there are evident square areas at the patch locations in saliency maps.
It means these patches have a high probability of being detected at people's first glance.
In contrast, adversarial patches generated by our approach can hardly be identified from saliency maps, which indicates that patches from our approach are relatively inconspicuous under human observations.

\subsubsection{User Evaluation}
In this section, we conduct a user study to evaluate the detection risks of adversarial patches.
We mainly compare our approach with Google Patch and \re{PS-GAN}.
Specifically, we prepare 100 images, in which 50 images are with IAP, and the rest are attached with a patch from \re{PS-GAN} or Google Patch or are simply natural images. 
These images are split into five Question Sets, and there are 20 images in each of the sets.
The participants are asked to pick up one question set and label all images on which they notice the existence of synthetic patches.
We remove \re{LaVAN} in this user evaluation for two reasons: one is that patches generated by \re{LaVAN} have no semantic information, which is similar to Google Patch; the other is to reduce the complexity of this evaluation and make it user friendly.
We collected 102 anonymous answer sheets online. 
We calculated the rates of images with patches that are labeled, and the statistics are summarized in Table~\ref{human identify}.
As we can see, patches produced by Google Patch and \re{PS-GAN} are of significantly higher probability to be detected while ours is much lower.
Though our approach cannot beat natural images on identified rates, it is far better than existing methods.
To summarize, our approach greatly reduces the detection risks. 

\begin{table}[t]
\caption{Average percentage of synthetic patches on the background images detected in user evaluation.}
\label{human identify}
\normalsize
\centering
\begin{tabular}{ccccc}
\toprule
Natural Image & Google Patch  & \re{PS-GAN}  & IAP \\ \midrule
12.15\%       & 93.63\%    & 89.90\% & 36.96\% \\ \bottomrule
\end{tabular}%
\end{table}

Further, we investigate whether the appearance of adversarial patches will influence the detection.
In our opinion, there are two main factors leading to an inconspicuous patch. 
Firstly, there should be sufficient consistency between patches and backgrounds. 
If there are high contrasts between patches and backgrounds, patches will be very likely to catch the user's visual attention.
Secondly, the appearance of patches should be natural; otherwise, the non-sense patterns will provide clues for observers.
Following the idea, we conduct a proof-of-concept experiment by asking participants to grade the patches' natural-looking appearances.
We prepare 120 images, of which 30 of them are the original image, 30 for the original image attached by Google Patch, 30 for \re{PS-GAN} patch, and 30 for IAP patch.
The best should be graded with $4$ and the worst graded with $1$ in each group of four approaches.
We collected 103 answer sheets, in which the users are at least of undergraduate education level. The average scores are shown in Table~\ref{human grades}.
As expected, our approach earns a much closer score to natural images.
What is more, we notice the correlation between Table~\ref{human identify} and Table~\ref{human grades}, which supports our claim that IAP is more inconspicuous and less detectable to observers compared to existing approaches.

\begin{table}[t]
\caption{Results of human evaluation grades on natural-looking appearance for each approach.}
\label{human grades}
\normalsize
\centering
\begin{tabular}{cccc}
\toprule
Natural Image & Google Patch & \re{PS-GAN} & IAP \\ \midrule
3.548         & 1.554        & 1.912  & 3.078   \\ \bottomrule
\end{tabular}%
\end{table}

\begin{figure}[t]
     \centering
     \begin{subfigure}[b]{0.23\columnwidth}
         \centering
         \includegraphics[width=\textwidth]{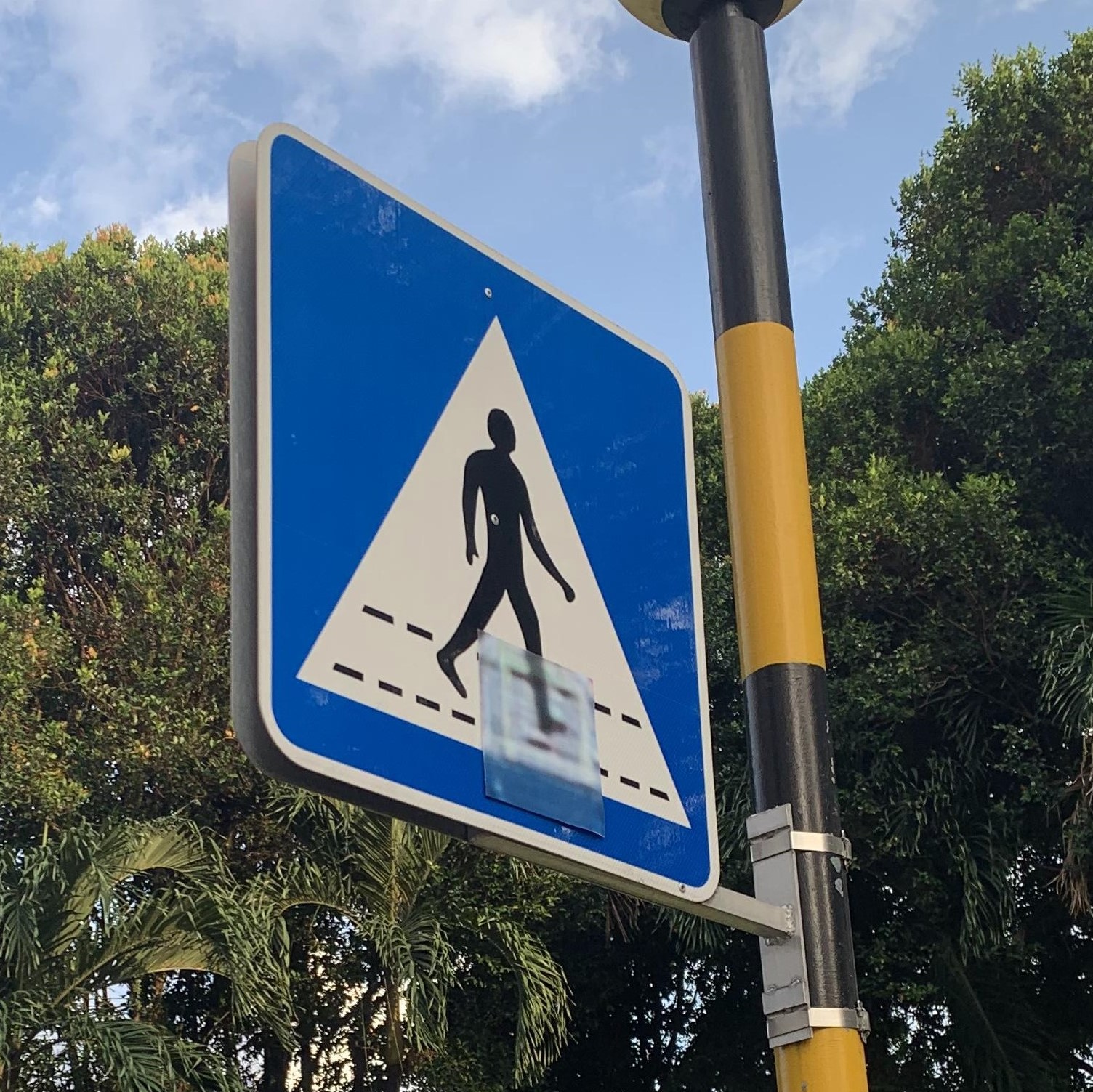}
         \caption{$-30\degree, 1m$}
         \label{fig:a}
     \end{subfigure}
    \hspace{0em}
     \begin{subfigure}[b]{0.23\columnwidth}
         \centering
        \includegraphics[width=\textwidth]{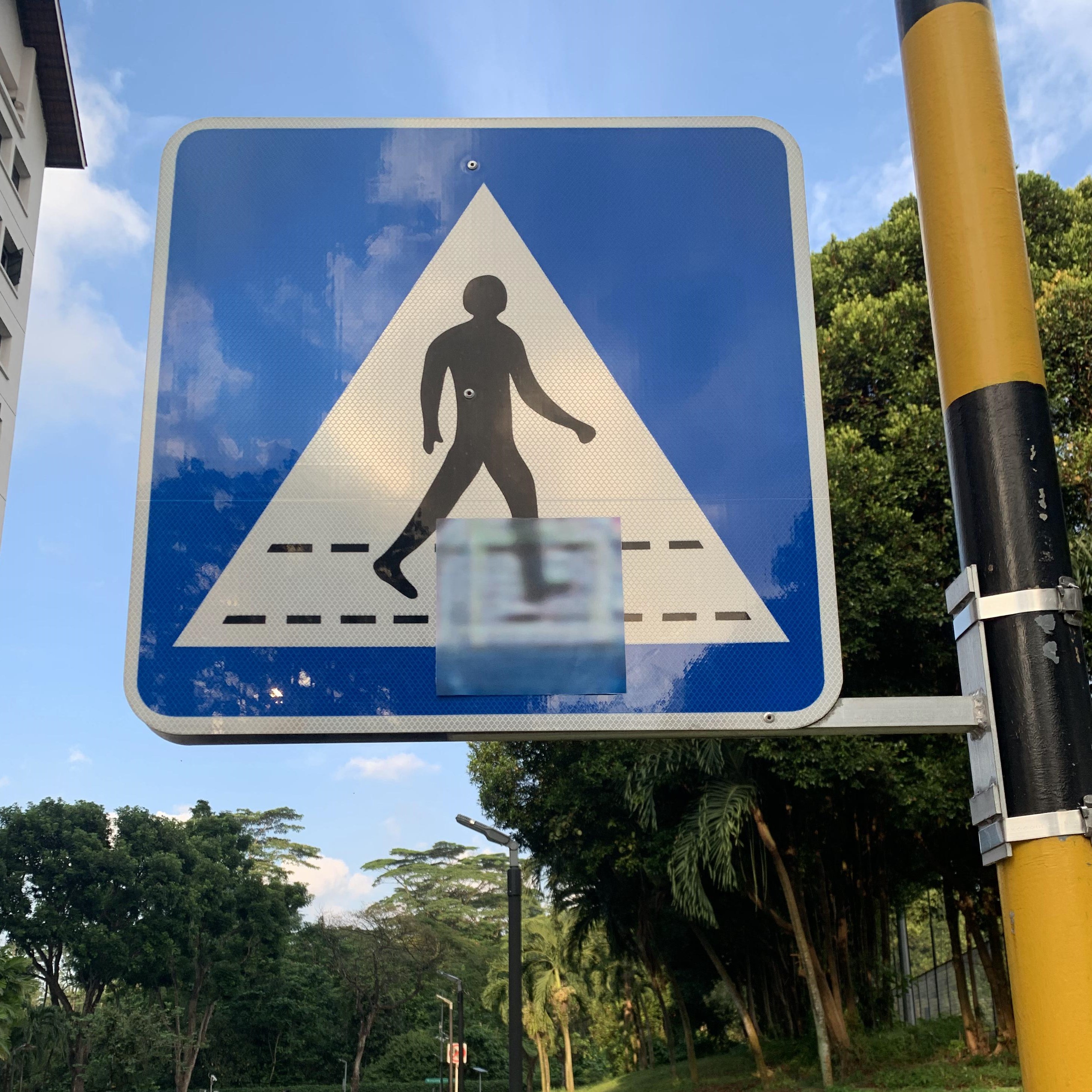}
         \caption{$0\degree, 1m$}
         \label{fig:b}
     \end{subfigure}
    \hspace{0em}
     \begin{subfigure}[b]{0.23\columnwidth}
         \centering
        \includegraphics[width=\textwidth]{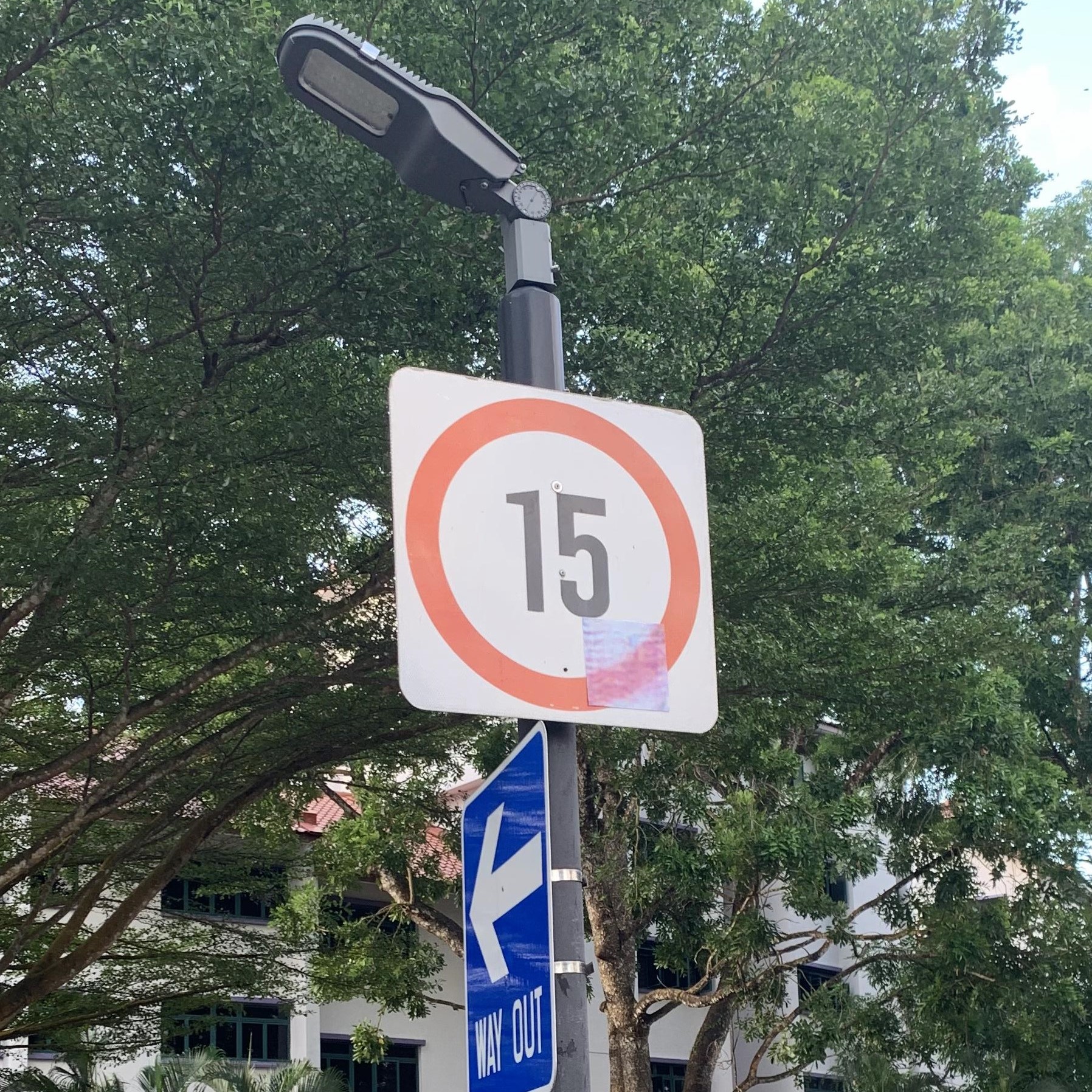}
         \caption{$-15\degree, 3m$}
         \label{fig:c}
     \end{subfigure}
     \hspace{0em}
     \begin{subfigure}[b]{0.23\columnwidth}
         \centering
         \includegraphics[width=\textwidth]{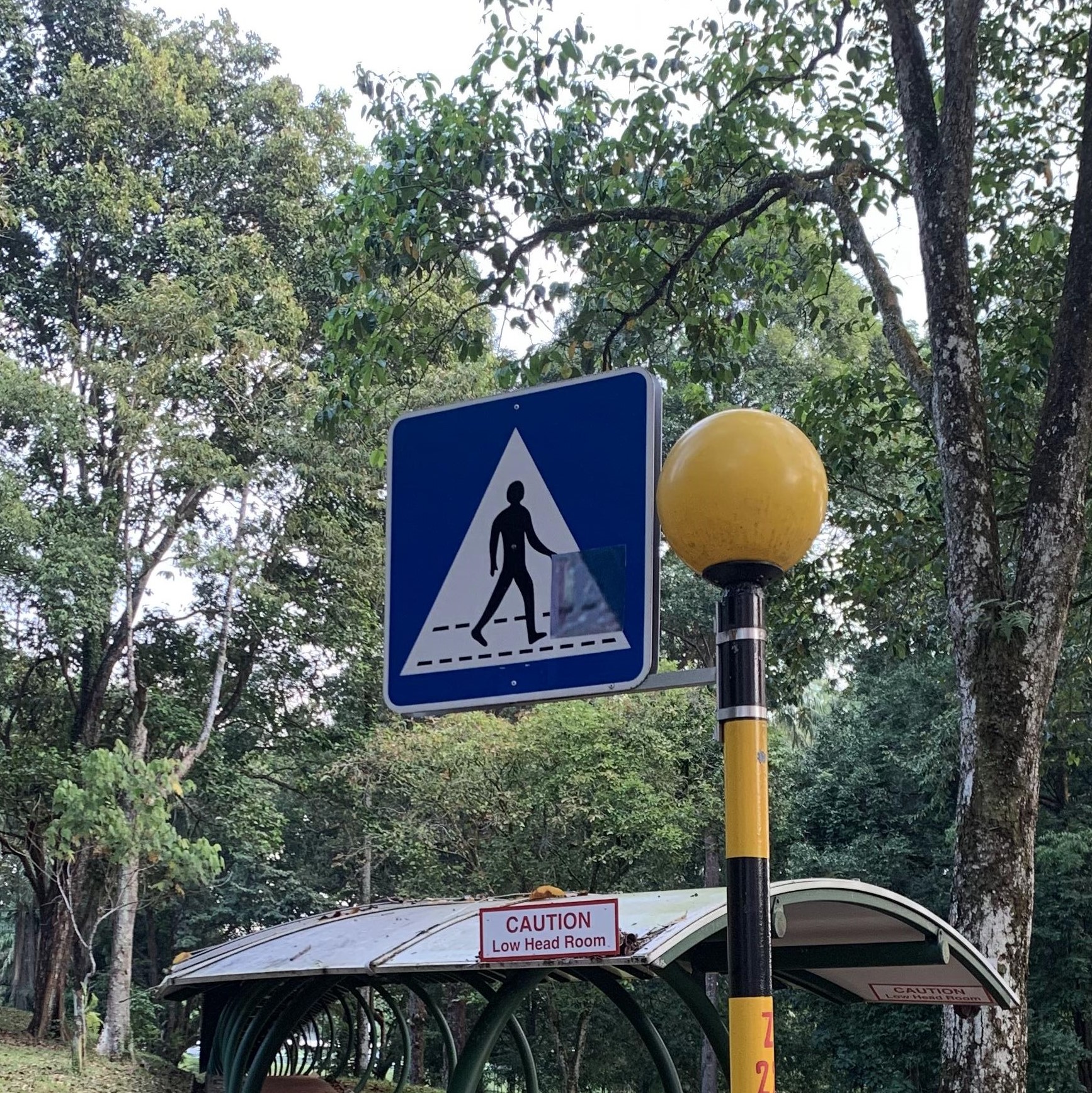}
         \caption{$30\degree, 3m$}
         \label{fig:d}
     \end{subfigure}
        \caption{Examples of our physical attack taken from four different places. From left to right, these traffic signs are classified as (a) 733 Pole, (b) 733 Pole, (c) 779 school bus, and (d) 971 bubble, respectively, which shows the effectiveness of our attack method in the physical world.}
        \label{real_world}
\end{figure}

\subsection{Physical Attack in Real World}
In this section, we conduct an experiment to show the attacking effectiveness of our approach in real-world settings. 
We choose five real-world traffic signboards on streets: three are Pedestrian-Ahead, one is Speed-Limit-15, and one is No-Entrance.
We first take one photo from each board by horizontal angles $\theta = \{0\degree\}$ and $d = \{1m\}$ (camera orthodox and closest to signs).
Here we use the pretrained InceptionV3\footnote{https://pytorch.org/hub/pytorch_vision_inception_v3/} as the target model, which correctly classifies these five traffic sign photos as ``920 Traffic Sign". 
Then we trained our IAP on these five images. 
For enhancing the real-world attack capability of IAP, we add the loss function (Equation~\ref{eqn:overall loss}) with a non-printability score~\cite{10.1145/2976749.2978392}
 \begin{equation}
 \begin{aligned}
& \mathcal{L}_{print} = \sum_{p'\in \tilde{p}}\space \prod_{a\in A}|p'-a|,
 \label{printable loss}
 \end{aligned}
 \end{equation}
where $A$ is the available collection of printable RGB values of $[0,1]^3$ and $p'$ is a pixel in our adversarial patch $\tilde{p}$. This is to make sure that the colors of our digital adversarial patches are printable from printers.
After the training was finished, we printed out the generated adversarial patches.
The model of printer we used is HP Color LaserJet CP3525dn. 
To ensure the fairness of natural lighting conditions, we attached printed patches on the signs correspondingly and took photos of each sign at the same time on another day. 

\re{Following the physical scenario settings in~\cite{Liu2019PerceptualSensitiveGF}, we use an iPhone Xs Max to take photos of five traffic signboards with various horizontal angles $\theta = \{-30\degree, -15\degree, 0\degree, 15\degree, 30\degree\}$ and distances $d = \{1m, 2m, 3m\}$.
In summary, we collect 75 pairs of clean samples and samples with adversarial patches.
With necessary preprocessing, these photos are fed into the target classifier.
The classification statistics show that 67 clean samples are classified correctly, while only 34 samples with our adversarial patches are classified correctly.
The accuracy of the target classifier is brought down by IAP from 89.3\% to 45.3\%. 
Similarly, we regenerated five adversarial patches for targeted attacks, where the target class is randomly selected as ``733 Pole".
The results show that 26 out of 75 samples with adversarial patches are predicted in class 733, the targeted success rate of which is 34.7\%.
Through the experiments in physical scenario, IAP is shown to be intriguing to craft strong adversarial attacks with limited real-world data.
Samples of our physical attacks are illustrated in Fig.~\ref{real_world}.
}

\subsection{Differences from adversarial perturbations}
\begin{figure}[t]
     \centering
     \begin{subfigure}[b]{0.48\columnwidth}
         \centering
         \includegraphics[width=\textwidth]{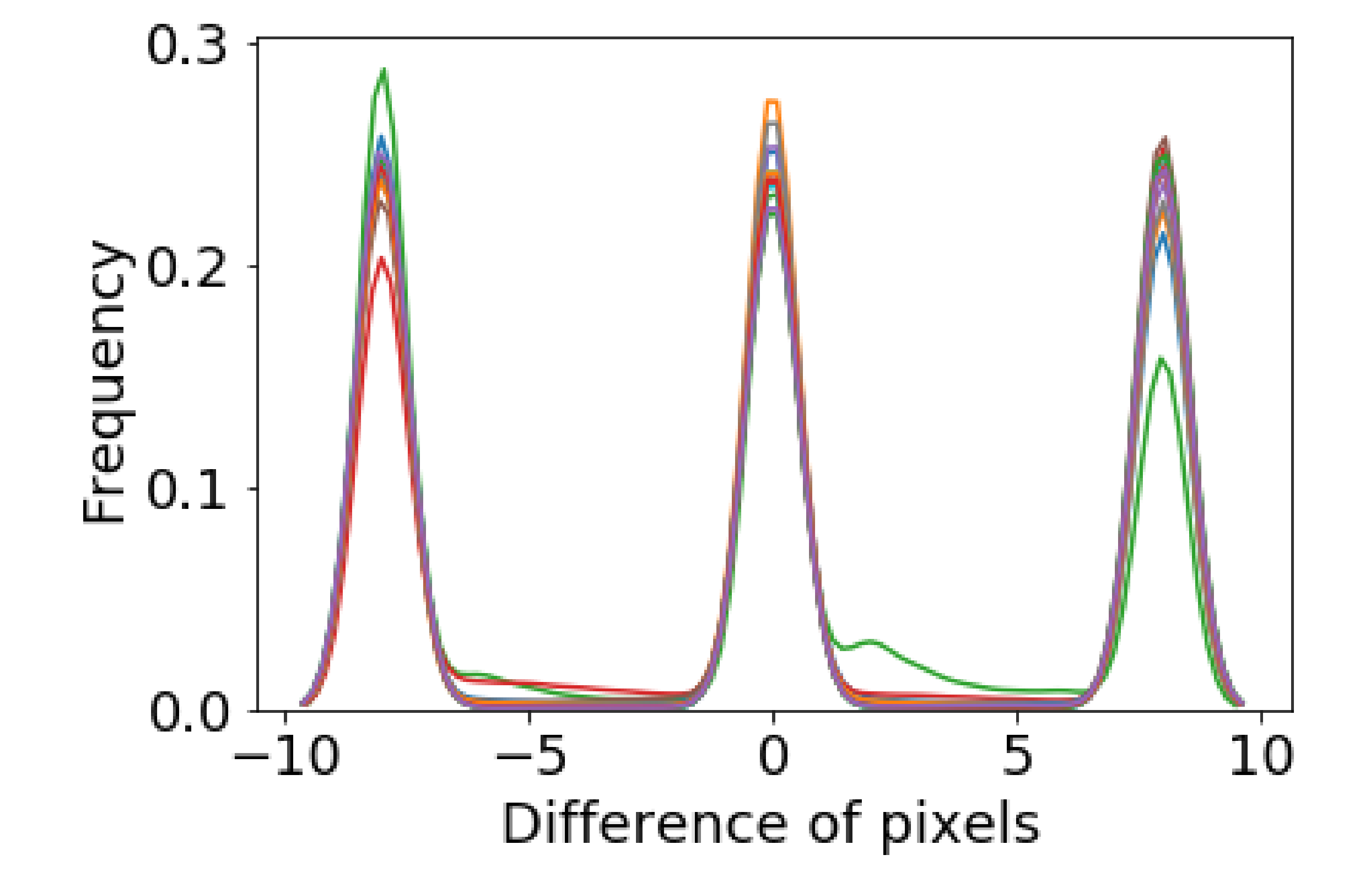}
         \caption{}
         \label{pgd_distribution}
     \end{subfigure}
     \hspace{0em}
     \begin{subfigure}[b]{0.48\columnwidth}
         \centering
         \includegraphics[width=\textwidth]{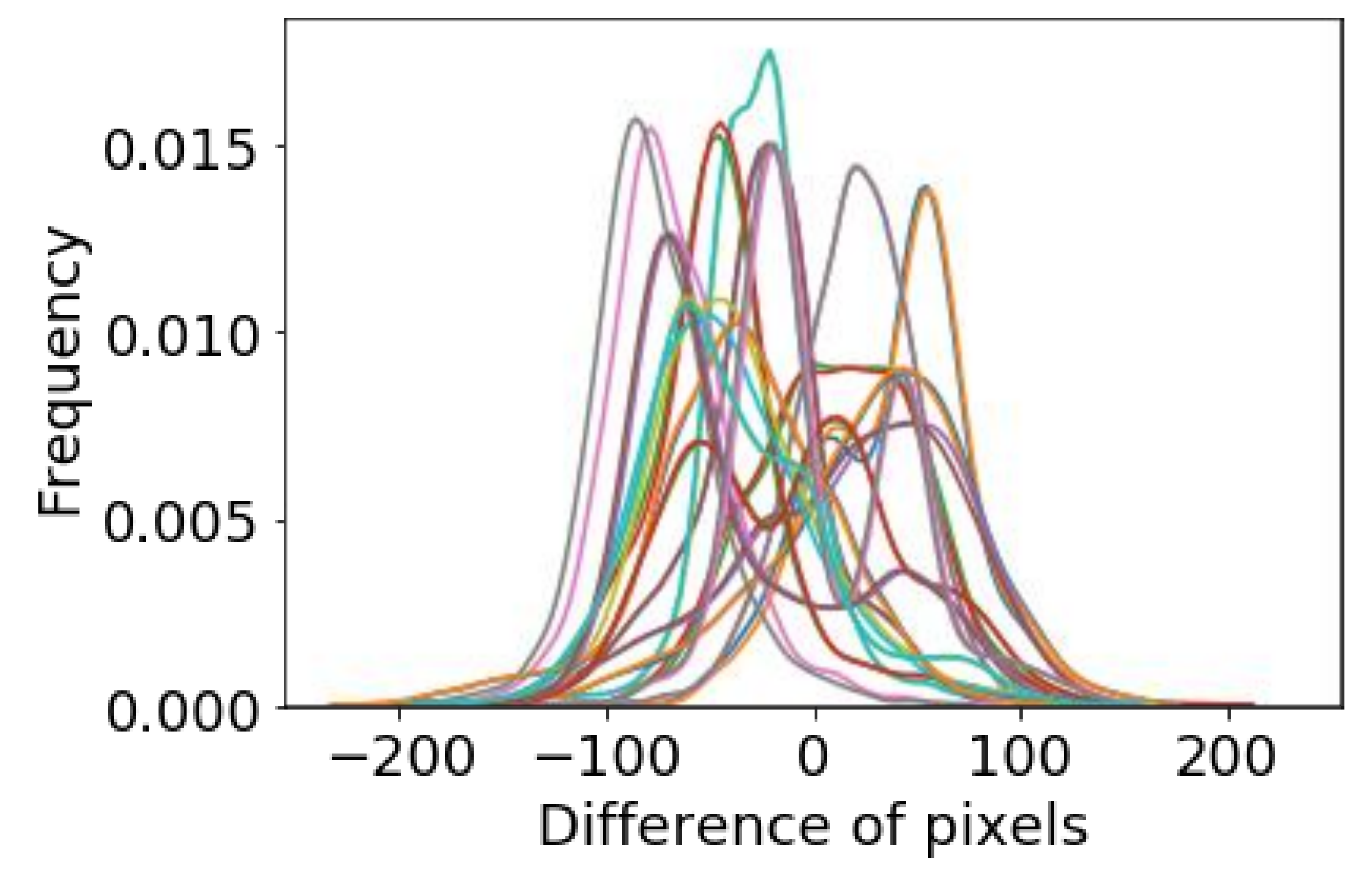}
         \caption{}
         \label{IAP_patch_distribution}
     \end{subfigure}
        \caption{Examples of value distribution obtained by (a) subtracting PGD noise ($\epsilon=8$) with original sample, and (b) subtracting IAP with the original patch.}
        \label{patch_distribution}
\end{figure}

This paper focuses on generating inconspicuous adversarial patches, which are designed to evade detection from humans.
Though images with our patches are naturally similar to the original images, they differ from adversarial perturbations.
Adversarial perturbations are strictly within a small $L_p$ norm ball, making them imperceptive to observers.
We randomly choose 15 target images and their corresponding IAPs generated by our approach to show the differences. 
We also perform PGD attack ($\epsilon=8$) \cite{madry2018towards} and obtain the adversarial images. 
Then we calculate the pixel value differences between original images and adversarial examples of PGD attacks and IAP.
We visualize the distributions of pixel value differences in Fig.~\ref{patch_distribution}. 
We can see that perturbations generated by PGD attacks are restricted within a given radius.
In contrast, the modification of pixel values on images by IAP roughly ranges from [-200, 200].
To conclude, our adversarial patches are quite different from adversarial perturbations, though both of which are non-observable.

\section{Conclusion}\label{sec:Conclusion}
This paper proposes an approach to generate inconspicuous adversarial patches for fooling image recognition systems on mobile devices.
With only one single image as the training data, IAP dramatically reduces the training cost.
Our approach shows the satisfying attack capabilities, black-box transferability, and deployable potential in real-world settings through extensive experiments.
Unlike existing works, we consider the consistency between patches and original images, and IAP is proved to be effective in reducing detection risks during qualitative and quantitative evaluations.
We address the potential applications of inconspicuous adversarial patches in real-world practices.
In the future, we will investigate developing adversarial patches with object awareness and make stronger attacks.

\ifCLASSOPTIONcaptionsoff
  \newpage
\fi

\bibliographystyle{IEEEtran}
\bibliography{bibliography}

\begin{thebibliography}{10}
\providecommand{\url}[1]{#1}
\csname url@samestyle\endcsname
\providecommand{\newblock}{\relax}
\providecommand{\bibinfo}[2]{#2}
\providecommand{\BIBentrySTDinterwordspacing}{\spaceskip=0pt\relax}
\providecommand{\BIBentryALTinterwordstretchfactor}{4}
\providecommand{\BIBentryALTinterwordspacing}{\spaceskip=\fontdimen2\font plus
\BIBentryALTinterwordstretchfactor\fontdimen3\font minus
  \fontdimen4\font\relax}
\providecommand{\BIBforeignlanguage}[2]{{%
\expandafter\ifx\csname l@#1\endcsname\relax
\typeout{** WARNING: IEEEtran.bst: No hyphenation pattern has been}%
\typeout{** loaded for the language `#1'. Using the pattern for}%
\typeout{** the default language instead.}%
\else
\language=\csname l@#1\endcsname
\fi
#2}}
\providecommand{\BIBdecl}{\relax}
\BIBdecl

\bibitem{Brown2017AdversarialP}
T.~B. Brown, D.~Man{\'e}, A.~Roy, M.~Abadi, and J.~Gilmer, ``Adversarial
  patch,'' \emph{ArXiv}, 2017.

\bibitem{karmon2018lavan}
D.~Karmon, D.~Zoran, and Y.~Goldberg, ``Lavan: Localized and visible
  adversarial noise,'' in \emph{International Conference on Machine
  Learning}.\hskip 1em plus 0.5em minus 0.4em\relax PMLR, 2018, pp. 2507--2515.

\bibitem{Liu2019PerceptualSensitiveGF}
A.~Liu, X.~Liu, J.~Fan, Y.~Ma, A.~Zhang, H.~Xie, and D.~Tao,
  ``Perceptual-sensitive {GAN} for generating adversarial patches,'' in
  \emph{The Thirty-Third {AAAI} Conference on Artificial Intelligence, {AAAI}
  2019, The Thirty-First Innovative Applications of Artificial Intelligence
  Conference, {IAAI} 2019, The Ninth {AAAI} Symposium on Educational Advances
  in Artificial Intelligence, {EAAI} 2019, Honolulu, Hawaii, USA, January 27 -
  February 1, 2019}, 2019, pp. 1028--1035.

\bibitem{zanella2014internet}
A.~Zanella, N.~Bui, A.~Castellani, L.~Vangelista, and M.~Zorzi, ``Internet of
  things for smart cities,'' \emph{IEEE Internet of Things journal}, no.~1, pp.
  22--32, 2014.

\bibitem{jin2014information}
J.~Jin, J.~Gubbi, S.~Marusic, and M.~Palaniswami, ``An information framework
  for creating a smart city through internet of things,'' \emph{IEEE Internet
  of Things journal}, no.~2, pp. 112--121, 2014.

\bibitem{owayjan2015face}
M.~Owayjan, A.~Dergham, G.~Haber, N.~Fakih, A.~Hamoush, and E.~Abdo, ``Face
  recognition security system,'' in \emph{New Trends in Networking, Computing,
  E-learning, Systems Sciences, and Engineering}, 2015, pp. 343--348.

\bibitem{bhowmik2011thermal}
M.~K. Bhowmik, K.~Saha, S.~Majumder, G.~Majumder, A.~Saha, A.~N. Sarma,
  D.~Bhattacharjee, D.~K. Basu, and M.~Nasipuri, ``Thermal infrared face
  recognition—a biometric identification technique for robust security
  system,'' \emph{Reviews, refinements and new ideas in face recognition},
  2011.

\bibitem{aria2020secure}
M.~Aria, M.~V. Agnihotri, M.~A. Rohra, and M.~R. Sekhar, ``Secure online
  payment with facial recognition using mtcnn,'' \emph{International Journal of
  Applied Engineering Research}, no.~3, pp. 249--252, 2020.

\bibitem{deng2020analysis}
Y.~Deng, X.~Zheng, T.~Zhang, C.~Chen, G.~Lou, and M.~Kim, ``An analysis of
  adversarial attacks and defenses on autonomous driving models,'' \emph{arXiv
  preprint arXiv:2002.02175}, 2020.

\bibitem{patricio2018computer}
D.~I. Patr{\'\i}cio and R.~Rieder, ``Computer vision and artificial
  intelligence in precision agriculture for grain crops: A systematic review,''
  \emph{Computers and electronics in agriculture}, pp. 69--81, 2018.

\bibitem{Szegedy2014IntriguingPO}
C.~Szegedy, W.~Zaremba, I.~Sutskever, J.~Bruna, D.~Erhan, I.~J. Goodfellow, and
  R.~Fergus, ``Intriguing properties of neural networks,'' in \emph{2nd
  International Conference on Learning Representations, {ICLR} 2014, Banff, AB,
  Canada, April 14-16, 2014, Conference Track Proceedings}, 2014.

\bibitem{MoosaviDezfooli2016DeepFoolAS}
S.~Moosavi{-}Dezfooli, A.~Fawzi, and P.~Frossard, ``Deepfool: {A} simple and
  accurate method to fool deep neural networks,'' in \emph{2016 {IEEE}
  Conference on Computer Vision and Pattern Recognition, {CVPR} 2016, Las
  Vegas, NV, USA, June 27-30, 2016}, 2016, pp. 2574--2582.

\bibitem{Carlini_2017}
N.~Carlini and D.~Wagner, ``Towards evaluating the robustness of neural
  networks,'' \emph{2017 IEEE Symposium on Security and Privacy (SP)}, 2017.

\bibitem{papernot2016limitations}
N.~Papernot, P.~McDaniel, S.~Jha, M.~Fredrikson, Z.~B. Celik, and A.~Swami,
  ``{The limitations of deep learning in adversarial settings},'' in \emph{2016
  IEEE European Symposium on Security and Privacy (EuroS{\&}P)}.\hskip 1em plus
  0.5em minus 0.4em\relax IEEE, 2016, pp. 372--387.

\bibitem{su2019one}
J.~Su, D.~V. Vargas, and K.~Sakurai, ``{One pixel attack for fooling deep
  neural networks},'' \emph{IEEE Transactions on Evolutionary Computation},
  2019.

\bibitem{Liu_2019_ICCV}
H.~Liu, R.~Ji, J.~Li, B.~Zhang, Y.~Gao, Y.~Wu, and F.~Huang, ``Universal
  adversarial perturbation via prior driven uncertainty approximation,'' in
  \emph{2019 {IEEE/CVF} International Conference on Computer Vision, {ICCV}
  2019, Seoul, Korea (South), October 27 - November 2, 2019}, 2019, pp.
  2941--2949.

\bibitem{ijcai2019-134}
X.~Wei, S.~Liang, N.~Chen, and X.~Cao, ``Transferable adversarial attacks for
  image and video object detection,'' in \emph{Proceedings of the Twenty-Eighth
  International Joint Conference on Artificial Intelligence, {IJCAI} 2019,
  Macao, China, August 10-16, 2019}, 2019, pp. 954--960.

\bibitem{yan2020cooling}
B.~Yan, D.~Wang, H.~Lu, and X.~Yang, ``Cooling-shrinking attack: Blinding the
  tracker with imperceptible noises,'' in \emph{2020 {IEEE/CVF} Conference on
  Computer Vision and Pattern Recognition, {CVPR} 2020, Seattle, WA, USA, June
  13-19, 2020}, 2020, pp. 987--996.

\bibitem{arnab2018robustness}
A.~Arnab, O.~Miksik, and P.~H.~S. Torr, ``On the robustness of semantic
  segmentation models to adversarial attacks,'' in \emph{2018 {IEEE} Conference
  on Computer Vision and Pattern Recognition, {CVPR} 2018, Salt Lake City, UT,
  USA, June 18-22, 2018}, 2018, pp. 888--897.

\bibitem{DBLP:journals/corr/KurakinGB16}
A.~Kurakin, I.~J. Goodfellow, and S.~Bengio, ``{Adversarial examples in the
  physical world},'' \emph{CoRR}, 2016.

\bibitem{Wiyatno_2019_ICCV}
R.~Wiyatno and A.~Xu, ``Physical adversarial textures that fool visual object
  tracking,'' in \emph{2019 {IEEE/CVF} International Conference on Computer
  Vision, {ICCV} 2019, Seoul, Korea (South), October 27 - November 2, 2019},
  2019, pp. 4821--4830.

\bibitem{DBLP:conf/icml/LiSK19}
J.~Li, F.~R. Schmidt, and J.~Z. Kolter, ``Adversarial camera stickers: {A}
  physical camera-based attack on deep learning systems,'' in \emph{Proceedings
  of the 36th International Conference on Machine Learning, {ICML} 2019, 9-15
  June 2019, Long Beach, California, {USA}}, 2019, pp. 3896--3904.

\bibitem{Luo_Bai_Zhao_2021}
\BIBentryALTinterwordspacing
J.~Luo, T.~Bai, and J.~Zhao, ``{Generating Adversarial yet Inconspicuous
  Patches with a Single Image (Student Abstract)},'' \emph{Proceedings of the
  AAAI Conference on Artificial Intelligence}, vol.~35, no.~18, pp.
  15\,837--15\,838, may 2021. [Online]. Available:
  \url{https://ojs.aaai.org/index.php/AAAI/article/view/17915}
\BIBentrySTDinterwordspacing

\bibitem{43405}
I.~J. Goodfellow, J.~Shlens, and C.~Szegedy, ``Explaining and harnessing
  adversarial examples,'' in \emph{3rd International Conference on Learning
  Representations, {ICLR} 2015, San Diego, CA, USA, May 7-9, 2015, Conference
  Track Proceedings}, 2015.

\bibitem{Tramr2018EnsembleAT}
F.~Tram{\`{e}}r, A.~Kurakin, N.~Papernot, I.~J. Goodfellow, D.~Boneh, and P.~D.
  McDaniel, ``Ensemble adversarial training: Attacks and defenses,'' in
  \emph{6th International Conference on Learning Representations, {ICLR} 2018,
  Vancouver, BC, Canada, April 30 - May 3, 2018, Conference Track Proceedings},
  2018.

\bibitem{Kurakin2017AdversarialEI}
A.~Kurakin, I.~J. Goodfellow, and S.~Bengio, ``Adversarial examples in the
  physical world,'' \emph{ArXiv}, 2017.

\bibitem{madry2018towards}
A.~Madry, A.~Makelov, L.~Schmidt, D.~Tsipras, and A.~Vladu, ``Towards deep
  learning models resistant to adversarial attacks,'' in \emph{6th
  International Conference on Learning Representations, {ICLR} 2018, Vancouver,
  BC, Canada, April 30 - May 3, 2018, Conference Track Proceedings}, 2018.

\bibitem{9165820}
Y.~Li, X.~Xu, J.~Xiao, S.~Li, and H.~T. Shen, ``{Adaptive Square Attack:
  Fooling Autonomous Cars with Adversarial Traffic Signs},'' \emph{IEEE
  Internet of Things Journal}, p.~1, 2020.

\bibitem{inproceedings2018patch}
K.~Eykholt, I.~Evtimov, E.~Fernandes, B.~Li, A.~Rahmati, C.~Xiao, A.~Prakash,
  T.~Kohno, and D.~Song, ``Robust physical-world attacks on deep learning
  visual classification,'' in \emph{2018 {IEEE} Conference on Computer Vision
  and Pattern Recognition, {CVPR} 2018, Salt Lake City, UT, USA, June 18-22,
  2018}, 2018, pp. 1625--1634.

\bibitem{Liu2019DPATCHAA}
X.~Liu, H.~Yang, Z.~Liu, L.~Song, Y.~Chen, and H.~Li, ``Dpatch: An adversarial
  patch attack on object detectors,'' \emph{arXiv: Computer Vision and Pattern
  Recognition}, 2019.

\bibitem{Thys_2019}
S.~Thys, W.~V. Ranst, and T.~Goedeme, ``Fooling automated surveillance cameras:
  Adversarial patches to attack person detection,'' \emph{2019 IEEE/CVF
  Conference on Computer Vision and Pattern Recognition Workshops (CVPRW)},
  2019.

\bibitem{Lee2019OnPA}
M.~Lee and J.~Z. Kolter, ``On physical adversarial patches for object
  detection,'' \emph{ArXiv}, 2019.

\bibitem{10.5555/2969033.2969125}
I.~J. Goodfellow, J.~Pouget{-}Abadie, M.~Mirza, B.~Xu, D.~Warde{-}Farley,
  S.~Ozair, A.~C. Courville, and Y.~Bengio, ``Generative adversarial nets,'' in
  \emph{Advances in Neural Information Processing Systems 27: Annual Conference
  on Neural Information Processing Systems 2014, December 8-13 2014, Montreal,
  Quebec, Canada}, 2014, pp. 2672--2680.

\bibitem{jia2020adv}
X.~Jia, X.~Wei, X.~Cao, and X.~Han, ``Adv-watermark: {A} novel watermark
  perturbation for adversarial examples,'' in \emph{{MM} '20: The 28th {ACM}
  International Conference on Multimedia, Virtual Event / Seattle, WA, USA,
  October 12-16, 2020}, 2020, pp. 1579--1587.

\bibitem{NIPS2018_8052}
Y.~Song, R.~Shu, N.~Kushman, and S.~Ermon, ``Constructing unrestricted
  adversarial examples with generative models,'' in \emph{Advances in Neural
  Information Processing Systems 31: Annual Conference on Neural Information
  Processing Systems 2018, NeurIPS 2018, December 3-8, 2018, Montr{\'{e}}al,
  Canada}, 2018, pp. 8322--8333.

\bibitem{qiu2019semanticadv}
H.~Qiu, C.~Xiao, L.~Yang, X.~Yan, H.~Lee, and B.~Li, ``Semanticadv: Generating
  adversarial examples via attribute-conditional image editing,'' \emph{arXiv
  preprint arXiv:1906.07927}, 2019.

\bibitem{lu2017safetynet}
J.~Lu, T.~Issaranon, and D.~A. Forsyth, ``Safetynet: Detecting and rejecting
  adversarial examples robustly,'' in \emph{{IEEE} International Conference on
  Computer Vision, {ICCV} 2017, Venice, Italy, October 22-29, 2017}, 2017, pp.
  446--454.

\bibitem{metzen2017detecting}
J.~H. Metzen, T.~Genewein, V.~Fischer, and B.~Bischoff, ``On detecting
  adversarial perturbations,'' in \emph{5th International Conference on
  Learning Representations, {ICLR} 2017, Toulon, France, April 24-26, 2017,
  Conference Track Proceedings}, 2017.

\bibitem{meng2017magnet}
D.~Meng and H.~Chen, ``Magnet: a two-pronged defense against adversarial
  examples,'' in \emph{Proceedings of the 2017 ACM SIGSAC conference on
  computer and communications security}, 2017, pp. 135--147.

\bibitem{li2017adversarial}
X.~Li and F.~Li, ``Adversarial examples detection in deep networks with
  convolutional filter statistics,'' in \emph{{IEEE} International Conference
  on Computer Vision, {ICCV} 2017, Venice, Italy, October 22-29, 2017}, 2017,
  pp. 5775--5783.

\bibitem{xu2017feature}
W.~Xu, D.~Evans, and Y.~Qi, ``Feature squeezing: Detecting adversarial examples
  in deep neural networks,'' \emph{arXiv preprint arXiv:1704.01155}, 2017.

\bibitem{DBLP:journals/corr/SelvarajuDVCPB16}
R.~R. Selvaraju, A.~Das, R.~Vedantam, M.~Cogswell, D.~Parikh, and D.~Batra,
  ``Grad-cam: Why did you say that? visual explanations from deep networks via
  gradient-based localization,'' \emph{CoRR}, 2016.

\bibitem{li2016precomputed}
C.~Li and M.~Wand, ``Precomputed real-time texture synthesis with markovian
  generative adversarial networks,'' in \emph{European conference on computer
  vision}.\hskip 1em plus 0.5em minus 0.4em\relax Springer, 2016, pp. 702--716.

\bibitem{Isola_2017}
P.~Isola, J.~Zhu, T.~Zhou, and A.~A. Efros, ``Image-to-image translation with
  conditional adversarial networks,'' in \emph{2017 {IEEE} Conference on
  Computer Vision and Pattern Recognition, {CVPR} 2017, Honolulu, HI, USA, July
  21-26, 2017}, 2017, pp. 5967--5976.

\bibitem{zhu2017unpaired}
J.~Zhu, T.~Park, P.~Isola, and A.~A. Efros, ``Unpaired image-to-image
  translation using cycle-consistent adversarial networks,'' in \emph{{IEEE}
  International Conference on Computer Vision, {ICCV} 2017, Venice, Italy,
  October 22-29, 2017}, 2017, pp. 2242--2251.

\bibitem{Shaham_2019}
T.~R. Shaham, T.~Dekel, and T.~Michaeli, ``Singan: Learning a generative model
  from a single natural image,'' in \emph{2019 {IEEE/CVF} International
  Conference on Computer Vision, {ICCV} 2019, Seoul, Korea (South), October 27
  - November 2, 2019}, 2019, pp. 4569--4579.

\bibitem{gulrajani2017improved}
I.~Gulrajani, F.~Ahmed, M.~Arjovsky, V.~Dumoulin, and A.~C. Courville,
  ``Improved training of wasserstein gans,'' in \emph{Advances in Neural
  Information Processing Systems 30: Annual Conference on Neural Information
  Processing Systems 2017, December 4-9, 2017, Long Beach, CA, {USA}}, 2017,
  pp. 5767--5777.

\bibitem{imagenet_cvpr09}
J.~Deng, W.~Dong, R.~Socher, L.~Li, K.~Li, and F.~Li, ``Imagenet: {A}
  large-scale hierarchical image database,'' in \emph{2009 {IEEE} Computer
  Society Conference on Computer Vision and Pattern Recognition {(CVPR} 2009),
  20-25 June 2009, Miami, Florida, {USA}}, 2009, pp. 248--255.

\bibitem{DBLP:journals/corr/SzegedyVISW15}
C.~Szegedy, V.~Vanhoucke, S.~Ioffe, J.~Shlens, and Z.~Wojna, ``Rethinking the
  inception architecture for computer vision,'' in \emph{2016 {IEEE} Conference
  on Computer Vision and Pattern Recognition, {CVPR} 2016, Las Vegas, NV, USA,
  June 27-30, 2016}, 2016, pp. 2818--2826.

\bibitem{Szegedy_2015}
C.~Szegedy, W.~Liu, Y.~Jia, P.~Sermanet, S.~E. Reed, D.~Anguelov, D.~Erhan,
  V.~Vanhoucke, and A.~Rabinovich, ``Going deeper with convolutions,'' in
  \emph{{IEEE} Conference on Computer Vision and Pattern Recognition, {CVPR}
  2015, Boston, MA, USA, June 7-12, 2015}, 2015, pp. 1--9.

\bibitem{Tan_2019}
M.~Tan, B.~Chen, R.~Pang, V.~Vasudevan, M.~Sandler, A.~Howard, and Q.~V. Le,
  ``Mnasnet: Platform-aware neural architecture search for mobile,'' in
  \emph{{IEEE} Conference on Computer Vision and Pattern Recognition, {CVPR}
  2019, Long Beach, CA, USA, June 16-20, 2019}, 2019, pp. 2820--2828.

\bibitem{Sandler_2018}
M.~Sandler, A.~G. Howard, M.~Zhu, A.~Zhmoginov, and L.~Chen, ``Mobilenetv2:
  Inverted residuals and linear bottlenecks,'' in \emph{2018 {IEEE} Conference
  on Computer Vision and Pattern Recognition, {CVPR} 2018, Salt Lake City, UT,
  USA, June 18-22, 2018}, 2018, pp. 4510--4520.

\bibitem{Montabone2010HumanDU}
S.~Montabone and A.~Soto, ``Human detection using a mobile platform and novel
  features derived from a visual saliency mechanism,'' \emph{Image Vis.
  Comput.}, 2010.

\bibitem{10.1145/2976749.2978392}
M.~Sharif, S.~Bhagavatula, L.~Bauer, and M.~K. Reiter, ``Accessorize to a
  crime: Real and stealthy attacks on state-of-the-art face recognition,'' in
  \emph{Proceedings of the 2016 ACM SIGSAC Conference on Computer and
  Communications Security}, 2016.

\end{thebibliography}

\end{document}